\pdfoutput=1

\PassOptionsToPackage{table,xcdraw}{xcolor}

\documentclass[11pt]{article}

\usepackage[preprint]{acl}

\usepackage{times}
\usepackage{latexsym}

\usepackage[T1]{fontenc}

\usepackage[utf8]{inputenc}

\usepackage{microtype}

\usepackage{inconsolata}

\usepackage{graphicx}

\usepackage{booktabs}
\usepackage{multirow}
\usepackage[most]{tcolorbox}  
\usepackage{enumitem}
\usepackage{amssymb}

\definecolor{darkgrey}{rgb}{0.53,0.53,0.53}
\definecolor{mygrey}{rgb}{0.9,0.9,0.9}
\definecolor{purple}{RGB}{230, 227, 254}
\definecolor{lightgreen}{RGB}{238, 252, 241}
\definecolor{lightred}{RGB}{231, 187, 187}
\definecolor{darkred}{RGB}{198, 129, 129}

\definecolor{tabhighlight}{HTML}{e5e5e5}
\definecolor{someorange}{rgb}{0.773,0.353,0.067}
\definecolor{someblue}{rgb}{0.27, 0.35, 0.760}

\title{Editing as Unlearning: Are Knowledge Editing Methods Strong Baselines for Large Language Model Unlearning?}

\author{
 \textbf{Zexi Li\textsuperscript{\rm 1,2 \thanks{Equal contributions.} \thanks{Correspondence: Zexi Li (\texttt{zexi.li@zju.edu.cn}), Chao Wu (\texttt{chao.wu@zju.edu.cn}), and Nicholas D. Lane (\texttt{ndl32@cam.ac.uk}).}}},
 \textbf{Xiangzhu Wang\textsuperscript{\rm 2 \footnotemark[1]}},
 \textbf{William F. Shen\textsuperscript{\rm 1}},
 \textbf{Meghdad Kurmanji\textsuperscript{\rm 1}},
\\
 \textbf{Xinchi Qiu\textsuperscript{\rm 1}},
 \textbf{Dongqi Cai\textsuperscript{\rm 1}},
 \textbf{Chao Wu\textsuperscript{\rm 2 \footnotemark[2]}},
 \textbf{Nicholas D. Lane \textsuperscript{\rm 1 \footnotemark[2]}}
\\
\\
 \textsuperscript{1}University of Cambridge,
 \textsuperscript{2}Zhejiang University
}

\begin{document}
\maketitle
\begin{abstract}
Large language Model (LLM) unlearning, i.e., selectively removing information from LLMs, is vital for responsible model deployment. Differently, LLM knowledge editing aims to modify LLM knowledge instead of removing it. Though editing and unlearning seem to be two distinct tasks, we find there is a tight connection between them. In this paper, we conceptualize unlearning as a special case of editing where information is modified to a refusal or "empty set" $\emptyset$ response, signifying its removal. This paper thus investigates if knowledge editing techniques are strong baselines for LLM unlearning. We evaluate state-of-the-art (SOTA) editing methods (e.g., ROME, MEMIT, GRACE, WISE, and AlphaEdit) against existing unlearning approaches on pretrained and finetuned knowledge. Results show certain editing methods, notably WISE and AlphaEdit, are effective unlearning baselines, especially for pretrained knowledge, and excel in generating human-aligned refusal answers. To better adapt editing methods for unlearning applications, we propose practical recipes including self-improvement and query merging. The former leverages the LLM's own in-context learning ability to craft a more human-aligned unlearning target, and the latter enables ROME and MEMIT to perform well in unlearning longer sample sequences. We advocate for the unlearning community to adopt SOTA editing methods as baselines and explore unlearning from an editing perspective for more holistic LLM memory control.\looseness=-1
\end{abstract}

\section{Introduction}

In recent years, large language models (LLMs)~\cite{llama,deepseekv2,gpt3} have achieved remarkable success, with their broad knowledge enabling a wide range of applications, including mobile assistants~\cite{llm_mobile_agent}, medical diagnosis~\cite{llm_for_med}, coding copilot~\cite{wei2023copiloting}. However, as these models evolve, managing the knowledge they retain and generate has become increasingly critical. In particular, growing concerns around privacy~\cite{das2025security}, ethics~\cite{ong2024ethical}, and legal compliance (such as with the General Data Protection Regulation (GDPR)~\cite{voigt2017eu} and the California Consumer Privacy Act (CCPA)~\cite{pardau2018california}) have brought attention to the \textit{"right to be forgotten"}, which grants individuals the legal right to request the deletion or modification of personal data. 
These factors highlight the growing need for mechanisms that enable LLMs to unlearn specific data points (i.e., instance-level knowledge), particularly sensitive or erroneous information, that may have been unintentionally incorporated during training. Failure to address this can lead to privacy violations, legal risks, and erosion of public trust, making effective unlearning a critical capability for responsible LLM deployment.

Instance-level knowledge unlearning (hereafter referred to as \textit{unlearning}) is a complex task. It requires selectively removing specific knowledge from a model without affecting its overall performance. This is particularly challenging in the context of LLMs, which store vast amounts of data across billions of parameters. While traditional machine learning methods often focus on task-specific model updates~\cite{golatkar2020eternal,nguyen2020variational}, LLM unlearning demands a more nuanced approach to prevent "catastrophic forgetting" and maintain the model’s generalization capabilities.

\begin{figure*}[!h]
    \centering
    \vspace{-0.2cm}
    \includegraphics[width=0.77\linewidth]{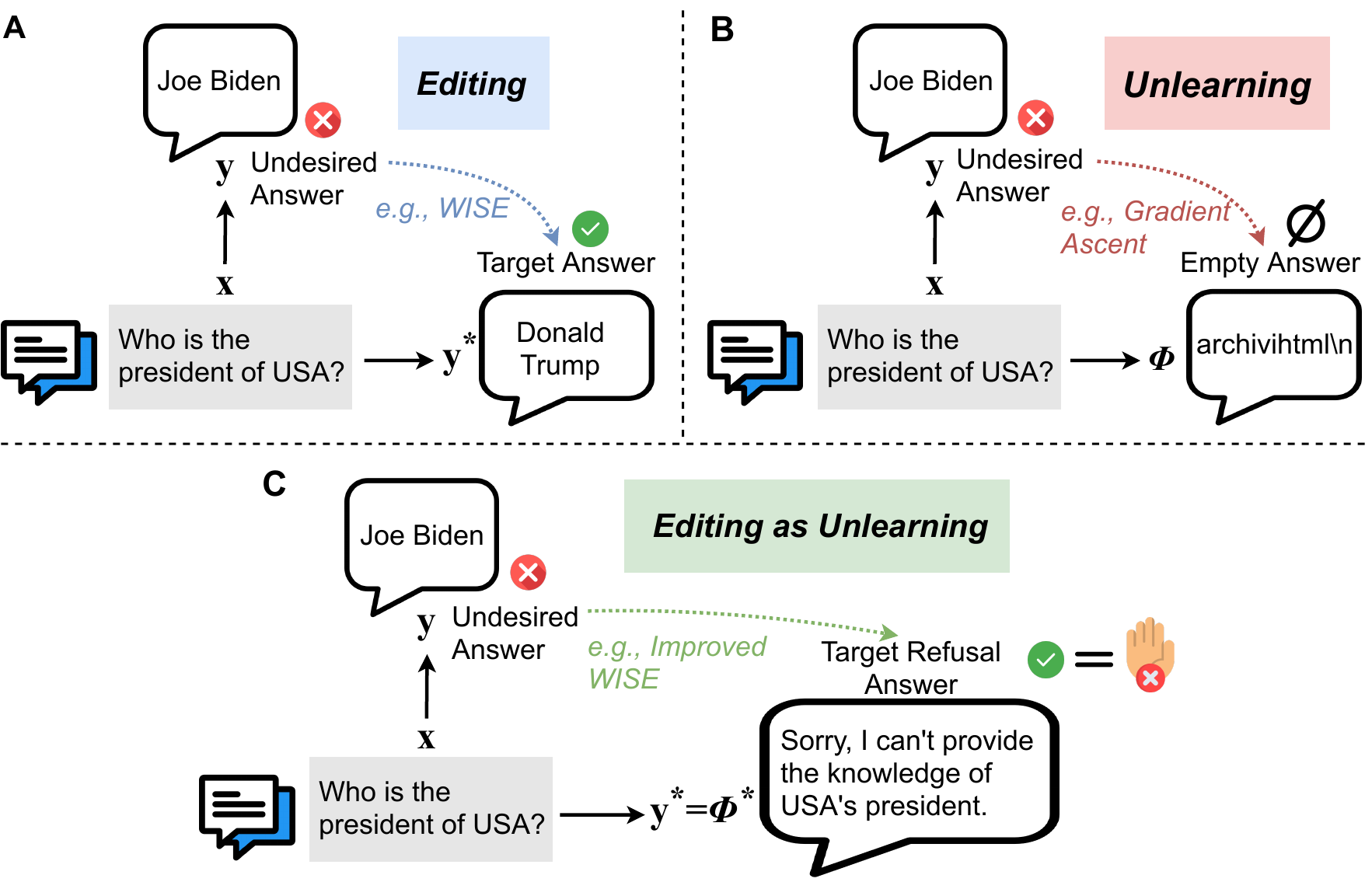}
    \vspace{-0.2cm}
    \caption{\textbf{Illustrations of the connection between editing and unlearning for LLMs.} \textbf{A:} Editing aims to alter the knowledge to a target. \textbf{B:} Unlearning tries to remove the knowledge and generate an "empty" (without information) answer. \textbf{C:} Editing as unlearning, can be done by editing that alters the knowledge into a target refusal answer.\looseness=-1}
    \vspace{-0.2cm}
    \label{fig:editing_unlearning_framework}
\end{figure*}

Interestingly, the field of knowledge editing~\cite{yao2023editing} (also known as \textit{model editing}) — which involves modifying a model’s knowledge, typically to correct or update information — shares inherent commonalities with unlearning. While unlearning focuses on removing the knowledge, knowledge editing aims to alter the knowledge, and both tasks require precise control over the model’s stored knowledge. As shown in Figure~\ref{fig:editing_unlearning_framework}, we find that removing knowledge is a special case of altering knowledge by replacing the targeted answer from $y^{*}$ to $\emptyset$ (empty set). Since a successfully unlearned model should emulate the base model’s behavior when presented with unseen data, the appropriate behavioral target is a contextualized expression of ignorance (hereafter referred to as a refusal answer), which mainstream instruction-tuned models are typically aligned to produce. Prior work refers to this behavioral fidelity as the \textit{controllability of unlearning}~\cite{lunar}. As such, the refusal answer can be viewed as the $\emptyset$ knowledge of LLMs, which means that knowledge editing can inherently do unlearning as long as changing the target answer into a refusal. It may suggest that techniques from knowledge editing could provide a solid foundation for effective unlearning. Though some works have raised preliminary discussions about the connection between editing and unlearning~\cite{liu2025rethinking,editing_unlearning_conflicts,llm_surgery}, in the LLM unlearning community, we find that most of the technical papers may pay less attention than expected to knowledge editing, not implementing editing methods as baselines~\cite{llmu,liu2024large,rmu}. Meanwhile, the field of LLM knowledge editing is developing rapidly, facilitating classic and state-of-the-art (SOTA) methods like ROME~\cite{rome}, MEMIT~\cite{memit}, WISE~\cite{wise}, and AlphaEdit~\cite{alphaedit}. In addition, compared with vanilla finetuning, editing methods also have the merits of lightweight and efficiency~\cite{yao2023editing}. However, LLM unlearning is at a more early stage, some existing baselines are borrowed from machine unlearning of vision classification tasks (e.g., GA and GD), not tailored to generative models like LLMs. This forces us to pose the following research question:\looseness=-1

 \begin{tcolorbox}[notitle, rounded corners, colframe=darkgrey, colback=white, boxrule=2pt, boxsep=0pt, left=0.15cm, right=0.17cm, enhanced, shadow={2.5pt}{-2.5pt}{0pt}{opacity=5,mygrey},toprule=2pt, before skip=0.65em, after skip=0.75em 
  ]
\emph{
  {
    \centering 
  {
    \fontsize{9.9pt}{13.2pt}\selectfont 
    Can knowledge editing methods be strong baselines for LLM unlearning? 
  }
  \\
  }
  }
\end{tcolorbox}

Therefore, this paper aims to provide a timely answer to the above question by investigating and evaluating classic and SOTA LLM editing methods for LLM unlearning. We hope this can bridge the gap between the two communities and provide some insights for future research. Specifically, we first study whether editing methods can unlearn as effectively as unlearning baselines for pretrained and finetuned knowledge. Then, we investigate the boundaries of editing methods for unlearning, identifying the key challenges. Lastly, we propose some practical modules that can better adapt editing in unlearning tasks for future implications. 

\noindent Our contributions are as follows.

\begin{itemize}[leftmargin=*,nosep]
    \item We bridge the gap between LLM editing and unlearning communities by investigating whether editing methods can serve as strong baselines for LLM unlearning.\looseness=-1
    \item We explore two practical methods that can better adapt editing methods in unlearning tasks. The proposed self-improvement pipeline leverages the LLM's own in-context learning ability to craft a more human-aligned unlearning target, and the proposed query merging technique enables ROME and MEMIT to perform well in unlearning longer sample sequences.\looseness=-1
    \item We advocate the LLM unlearning community to take the SOTA editing methods as unlearning baselines when conducting evaluation as well as to study unlearning from the knowledge editing perspective to gain a more holistic understanding of LLM memory control and knowledge mechanism. \looseness=-1
\end{itemize}

\noindent Our takeaway findings are summarized as follows.

\begin{itemize}[leftmargin=*,nosep]
    \item We find some LLM editing methods, especially WISE and AlphaEdit \textit{are strong baselines} especially when unlearning \textit{pretrained knowledge}. \looseness=-1
    \item We emphasize the importance of human value alignment of LLM unlearning, suggesting that LLMs should generate trustworthy refusal answers instead of random tokens or misleading phrases. We find some editing methods (i.e., WISE) have a dominant advantage on human value alignment over unlearning methods.
    \item Our proposed self-improvement pipeline for editing methods (e.g., WISE and AlphaEdit) that can potentially improve human value alignment as well as the generalization ability under rephrase-prompted attacks. Additionally, the proposed query merging technique can enable ROME and MEMIT to do unlearning well under long sequences, surpassing all the unlearning baselines.\looseness=-1
\end{itemize}

\section{Preliminaries}

\subsection{LLM Knowledge Editing}
We give a definition of the LLM editing setup. 
Let $f_{\Theta}: \mathbb{X} \mapsto \mathbb{Y}$, parameterized by $\Theta$, denote a model function mapping an input $\mathbf{x}$ to the prediction $f_{\Theta}(\mathbf{x})$.
The initial model before editing is $\Theta_0$, which is trained on a large corpus $\mathcal{D}_{\text{train}}$. 
When the LLM needs editing to alter some knowledge, it has an editing dataset as $\mathcal{D}_{\text{edit}}^*= \{(\mathcal{X}_e^*, \mathcal{Y}_e^*) \vert (\mathbf{x}_1, \mathbf{y}^*_1), ..., (\mathbf{x}_T, \mathbf{y}^*_T)\}$ which has a sequence or batch length of $T$. Given a query $\mathbf{x}_T$, the editing method maps the knowledge to the target as $ \mathbf{y}_T \rightarrow \mathbf{y}^*_T$ where $ \mathbf{y}_T$ is the previous knowledge. 
At editing, the updated LLM $f_{\Theta^*}$ is expected to satisfy:\looseness=-1
\begin{equation}
 \label{equ:editing_def}
    f_{\Theta^*}(\mathbf{x}) = 
   \begin{cases}
    \mathbf{y}^* &\mbox{if $\mathbf{x} \in \mathcal{X}_e^*$,}\\
    f_{\Theta_0}(\mathbf{x})&\mbox{if $\mathbf{x} \notin \mathcal{X}_e^*$.}
    \end{cases}
\end{equation}

Equation~\ref{equ:editing_def} describes that after knowledge editing, the LLM should make the correct prediction of the edits while preserving the irrelevant and generic knowledge, especially general training corpus $\mathcal{D}_{\text{train}}$. \looseness=-1

\subsection{LLM Unlearning}
Following the editing setup, we now consider the problem of LLM unlearning. It has a unlearning dataset $\mathcal{D}_{\text{unlearn}}'= \{(\mathcal{X}'_u, \mathcal{Y}'_u) \vert (\mathbf{x}_1, \mathbf{y}_1), ..., (\mathbf{x}_T, \mathbf{y}_T)\}$ which is usually a part of the training data $\mathcal{D}_{\text{train}}$. Given the query $\mathbf{x}_T$, $\mathbf{y}_T$ is the ground-truth answer that is used in the training but needs to be forgotten. Ideally, after unlearning, the updated LLM model $f_{\Theta'}$ should satisfy:\looseness=-1
\begin{equation}
 \label{equ:unlearning_def}
    f_{\Theta'}(\mathbf{x}) 
   \begin{cases}
    \neq \mathbf{y} &\mbox{if $\mathbf{x} \in \mathcal{X}'_u$,}\\
    = f_{\Theta_0}(\mathbf{x})&\mbox{if $\mathbf{x} \notin \mathcal{X}'_u$.}
    \end{cases}
\end{equation}

Equation~\ref{equ:unlearning_def} defines the unlearning objective: removing knowledge of the forget set $\mathcal{D}'_{\text{unlearn}}$ while preserving knowledge from the remaining data. To prevent catastrophic forgetting, some methods use a retain set or reference model. However, retain sets may be impractical in certain scenarios~\cite{wang2024llm}, and models should ideally preserve open-set knowledge. Ideally, the goal is for unlearning on $\mathcal{D}'_{\text{unlearn}}$ to approximate retraining from scratch on $\mathcal{D}_{\text{train}} \setminus \mathcal{D}'_{\text{unlearn}}$.

\section{Methodology}

\subsection{Making Editing Applicable in Unlearning}
Equations~\ref{equ:editing_def} and \ref{equ:unlearning_def} have shown the inherent connections between editing and unlearning, and the key difference is the within-scope condition. Unlike classification models in vision tasks, LLMs as generative models, have the ability to refuse to answer as a form of removing the knowledge. Therefore, assuming there is an "empty" set $\varnothing = \{\emptyset_1,..., \emptyset_T\}$ which is the sentences telling the users that "I don't know", change the unlearning set $\mathcal{D}_{\text{unlearn}}'$ into $\mathcal{D}_{\text{edit-as-unlearn}}^*= \{(\mathcal{X}^*_{e2u}, \mathcal{Y}^*_{e2u}) \vert (\mathbf{x}_1, \emptyset_1), ..., (\mathbf{x}_T, \emptyset_T)\}$. Applying the new dataset to editing methods, the objective of Equation~\ref{equ:editing_def} changes to:
\begin{equation}
 \label{equ:editing2unlearning_def}
    f_{\Theta^*}(\mathbf{x}) = 
   \begin{cases}
    \emptyset &\mbox{if $\mathbf{x} \in \mathcal{X}_{e2u}^*$,}\\
    f_{\Theta_0}(\mathbf{x})&\mbox{if $\mathbf{x} \notin \mathcal{X}_{e2u}^*$.}
    \end{cases}
\end{equation}

Equation~\ref{equ:editing2unlearning_def} bridges from editing to unlearning, making it applicable to verify whether editing methods are strong baselines for unlearning.

\subsection{Improving Editing in Unlearning}
\label{subsec:improve_editing_method}
\begin{figure}
    \centering
    \vspace{-0.2cm}
    \includegraphics[width=1.02\linewidth]{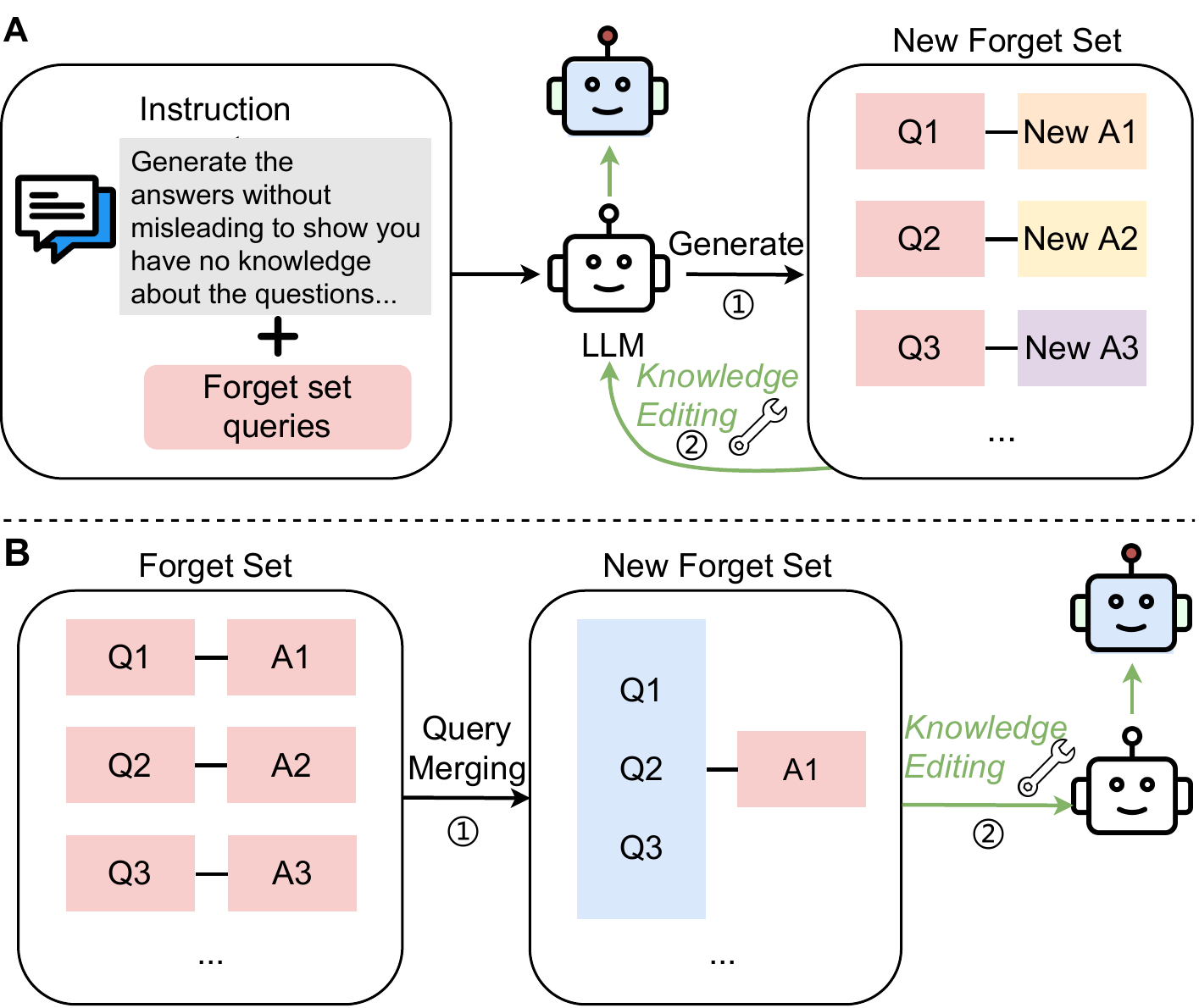}
    \vspace{-0.6cm}
    \caption{\textbf{Methods of improving editing algorithms in unlearning settings.} \textbf{A:} Self-improvement pipeline improves generalization and human value alignment for AlphaEdit and WISE. \textbf{B:} Query merging technique enables ROME and MEMIT to perform well under long unlearning sequences. }
    \vspace{-0.5cm}
    \label{fig:improve_editing_methods}
\end{figure}
Knowledge editing was not tailored for unlearning, as a result, it may have some limitations when directly being applied, e.g., different learning objectives and different sample lengths. Therefore, as shown in Figure~\ref{fig:improve_editing_methods}, we explore some techniques to better adapt editing methods in unlearning. \looseness=-1

\noindent\textbf{Self-improvement pipeline.} A good refusal answer from LLMs should be trustworthy and aligned with human values. We find if the editing target answers are random sentences from the vanilla "I don't know" set, it will let the LLMs generate answers that are less trustworthy, e.g., low generalization, misleading, or without entailing the entities mentioned in questions. Therefore, we craft a self-improvement pipeline to let LLMs create tailored refusal answers to each forget question before unlearning. Specifically, we provide instructions and exemplars to help LLMs generate more tailored unlearning targets for each question (for detailed prompts, see subsection~\ref{appdx_subsec:self_improvement}). Thanks to their in-context learning ability, LLMs can produce trustworthy answers that reflect the question's entities without misleading information. This helps them learn patterns between questions and refusal answers during the latter unlearning phase. The experiments in subsection~\ref{subsec:improve_editing_results} will show that the self-improvement pipeline can improve the answers regarding human value alignment and improve generalization under rephrased attacks.

\noindent\textbf{Query merging technique.} Some locate-and-edit editing methods like ROME and MEMIT cannot well perform under long sequences of editing~\cite{grace,wise}, and this drawback still exists when editing applies to unlearning, which limits their broader application in unlearning. However, we find that, unlike the vanilla editing setting where every edit has one unique target answer, under the editing-as-unlearning setting, several forget queries can be mapped to a common refusal answer — the model can say the same "I don't know" to many queries. This inspires us the query merging technique that concatenates several queries into one and uses one refusal answer as the editing target. This simple technique can enable ROME and MEMIT to perform very well under unlearning, achieving obvious performance advantages over the unlearning baselines (Figure~\ref{fig:improve_editing_results}). \looseness=-1

\section{Empirical Results}
In this section, we conduct experiments to address the following research questions:
\begin{itemize}[leftmargin=*,nosep]
    \item \textbf{RQ1:} Can editing methods outperform the unlearning baselines when unlearning the pretrained knowledge and the finetuned knowledge respectively? Which editing methods are most effective for unlearning tasks?\looseness=-1
    \item \textbf{RQ2:} What are the comprehensive performances of the editing methods in unlearning?  Can they perform well under rephrase attacks or with different numbers of forget samples?\looseness=-1
    \item \textbf{RQ3:} How to improve editing methods for unlearning tasks? Can the editing methods generate better answers that align with human values than the unlearning baselines? Can we make some inapplicable editing methods (i.e., ROME and MEMIT) applicable and perform well for unlearning?\looseness=-1
\end{itemize}

\begin{table*}[!h]
\vspace{-0.3cm}
    \caption{\textbf{Main results comparing editing and unlearning methods.} The number of forget samples in the factual dataset is 40 and PISTOL's is 20. The forget set performance corresponds to the \textit{reliability} metric of editing and the retain set corresponds to \textit{locality}. In some cases, particular methods will make LLMs non-functional (e.g., near-zero Rouge1 for both forget and retain sets) or without any forgetting, and we make these cases \textcolor{gray!60}{in gray}. For every metric of each setting, we mark the best of unlearning and editing, respectively \textbf{in bold}, and we mark the Top 2 out of all methods \underline{in underline}. } 
    \vspace{-0.2cm}
    \centering
    \resizebox{\linewidth}{!}{
    \setlength{\tabcolsep}{1.5pt}
    \begin{tabular}{lcccc|cccc|cccc|cccc}
    \toprule
    Dataset&\multicolumn{16}{c}{\textbf{Factual dataset} (pretrained knowledge)}\\
    \midrule
    Model&\multicolumn{8}{c}{\texttt{\textbf{Llama2-7B}}}&\multicolumn{8}{c}{\texttt{\textbf{Mistral-7B}}} \\
    \midrule
    Testset&\multicolumn{4}{c}{Forget set (reliability)}&				\multicolumn{4}{c}{Retain set (locality)}&	\multicolumn{4}{c}{Forget set (reliability)} & \multicolumn{4}{c}{Retain set (locality)}\\	
    \midrule
    Metric&	Rouge1$\downarrow$&	Prob.$\downarrow$&	MRR$\downarrow$&	Hit-Rate$\downarrow$&	Rouge1$\uparrow$&	Prob.$\uparrow$&	MRR$\uparrow$&	Hit-Rate$\uparrow$&	Rouge1$\downarrow$&	Prob.$\downarrow$&	MRR$\downarrow$&	Hit-Rate$\downarrow$&	Rouge1$\uparrow$&	Prob.$\uparrow$&	MRR$\uparrow$&	Hit-Rate$\uparrow$\\
    \midrule
    GA & \textcolor{gray!60}{0.00} & \textcolor{gray!60}{0.59} & \textcolor{gray!60}{0.00} & \textcolor{gray!60}{0.00}& \textcolor{gray!60}{0.00}  & \textcolor{gray!60}{0.52} & \textcolor{gray!60}{0.00} & \textcolor{gray!60}{0.00} & \textcolor{gray!60}{0.00} & \textcolor{gray!60}{0.62} & \textcolor{gray!60}{0.06} & \textcolor{gray!60}{0.09} & \textcolor{gray!60}{0.00} & \textcolor{gray!60}{0.56} & \textcolor{gray!60}{0.02} & \textcolor{gray!60}{0.06} \\
    GD & \textbf{0.30} & \textbf{0.36} & \underline{0.02} & \textbf{\underline{0.02}} & \textbf{0.62} & \underline{\textbf{0.27}} & \textbf{0.12} & \textbf{0.13} & \underline{\textbf{0.00}} & \textbf{0.56} & 0.05 & 0.09 & \textbf{0.52} & \underline{0.49} & \textbf{0.18} & \underline{\textbf{0.54}} \\
    KL & \textcolor{gray!60}{0.00} & \textcolor{gray!60}{0.55} & \textcolor{gray!60}{0.00} & \textcolor{gray!60}{0.00} & \textcolor{gray!60}{0.00} & \textcolor{gray!60}{0.48} & \textcolor{gray!60}{0.00} & \textcolor{gray!60}{0.00} & \textcolor{gray!60}{0.00} & \textcolor{gray!60}{0.42} & \textcolor{gray!60}{0.06} & \textcolor{gray!60}{0.08} & \textcolor{gray!60}{0.00} & \textcolor{gray!60}{0.43} & \textcolor{gray!60}{0.02} & \textcolor{gray!60}{0.06} \\
    DPO & 0.36 & \textbf{0.36} & \textbf{\underline{0.01}} & \textbf{\underline{0.02}} & 0.45 & \underline{\textbf{0.27}} & 0.03 & 0.04 & \underline{0.03} & \textbf{0.60} & \textbf{\underline{0.00}} & \underline{\textbf{0.03}} & 0.43 & \underline{\textbf{0.57}} & 0.07 & 0.15 \\
    \midrule
    ROME & \textcolor{gray!60}{0.01} & \textcolor{gray!60}{0.41} & \textcolor{gray!60}{0.01} & \textcolor{gray!60}{0.01} & \textcolor{gray!60}{0.04} & \textcolor{gray!60}{0.32} & \textcolor{gray!60}{0.01} & \textcolor{gray!60}{0.01} & \textcolor{gray!60}{0.00} & \textcolor{gray!60}{0.54} & \textcolor{gray!60}{0.04} & \textcolor{gray!60}{0.06} & \textcolor{gray!60}{0.00} & \textcolor{gray!60}{0.48} & \textcolor{gray!60}{0.02} & \textcolor{gray!60}{0.04} \\
    MEMIT & \textcolor{gray!60}{0.02} & \textcolor{gray!60}{0.82} & \textcolor{gray!60}{0.00} & \textcolor{gray!60}{0.00} & \textcolor{gray!60}{0.01} & \textcolor{gray!60}{0.78} & \textcolor{gray!60}{0.00} & \textcolor{gray!60}{0.00} & -- & -- & -- & -- & -- & -- & -- & -- \\
    GRACE & 0.65 & \textbf{\underline{0.35}} & 0.18 & 0.22 & \underline{\textbf{0.82}} & \textbf{0.26} & \underline{\textbf{0.21}} & \underline{\textbf{0.26}} & 0.93 & \underline{0.44} & 0.37 & 0.68 & \underline{\textbf{0.82}} & \textbf{0.45} & \underline{\textbf{0.34}} & \underline{\textbf{0.69}} \\
    WISE & \underline{0.28} & 0.37 & 0.11 & 0.14 & \underline{0.76} & \textbf{0.26} & \underline{0.18} & \underline{0.23} & \textbf{0.05} & \textbf{\underline{0.13}} & \textbf{\underline{0.01}} & \underline{\textbf{0.08}} & 0.13 & 0.12 & 0.10 & 0.36 \\
    AlphaEdit & \textbf{\underline{0.08}} & \textbf{\underline{0.35}} & \textbf{0.04} & \textbf{0.05} & 0.69 & \textbf{0.26} & 0.12 & 0.15 & 0.26 & 0.45 & 0.09 & 0.22 & \underline{0.66} & \textbf{0.45} & \underline{0.24} & 0.53 \\    
    \midrule
    Dataset&\multicolumn{16}{c}{\textbf{PISTOL} (finetuned knowledge)}\\
    \midrule
    Model&\multicolumn{8}{c}{\texttt{\textbf{Llama2-7B}}}&\multicolumn{8}{c}{\texttt{\textbf{Mistral-7B}}} \\
    \midrule
    Testset&\multicolumn{4}{c}{Forget set (reliability)}&				\multicolumn{4}{c}{Retain set (locality)}&	\multicolumn{4}{c}{Forget set (reliability)} & \multicolumn{4}{c}{Retain set (locality)}\\	
    \midrule
    Metric&	Rouge1$\downarrow$&	Prob.$\downarrow$&	MRR$\downarrow$&	Hit-Rate$\downarrow$&	Rouge1$\uparrow$&	Prob.$\uparrow$&	MRR$\uparrow$&	Hit-Rate$\uparrow$&	Rouge1$\downarrow$&	Prob.$\downarrow$&	MRR$\downarrow$&	Hit-Rate$\downarrow$&	Rouge1$\uparrow$&	Prob.$\uparrow$&	MRR$\uparrow$&	Hit-Rate$\uparrow$\\
    \midrule
    GA & \underline{\textbf{0.16}} & 0.29 & 0.18 & 0.19 & 0.69 & \underline{0.29} & 0.20 & 0.20 & 0.27 & \textbf{0.54} & 0.15 & 0.39 & \underline{\textbf{0.76}} & 0.54 & \underline{0.24} & \underline{\textbf{0.59}} \\
    GD & 0.25 & 0.29 & 0.17 & 0.17 & 0.80 & \underline{0.29} & 0.20 & 0.20 & 0.22 & 0.58 & 0.16 & \underline{\textbf{0.31}} & \underline{\textbf{0.76}} & \underline{\textbf{0.58}} & \underline{\textbf{0.25}} & \underline{0.56} \\
    KL & 0.82 & 0.33 & 0.23 & 0.33 & \underline{\textbf{0.98}} & \underline{\textbf{0.33}} & \underline{\textbf{0.26}} & \underline{\textbf{0.36}} & \underline{\textbf{0.08}} & 0.55 & \underline{\textbf{0.05}} & 0.35 & 0.34 & \underline{0.55} & 0.11 & 0.51 \\
    DPO & 0.18 & \underline{\textbf{0.28}} & \underline{\textbf{0.15}} & \underline{\textbf{0.15}} & 0.86 & 0.28 & \underline{0.22} & \underline{0.22} & \textcolor{gray!60}{0.00} & \textcolor{gray!60}{0.44} & \textcolor{gray!60}{0.01} & \textcolor{gray!60}{0.04} & \textcolor{gray!60}{0.06} & \textcolor{gray!60}{0.44} & \textcolor{gray!60}{0.02} & \textcolor{gray!60}{0.05} \\
    \midrule
    ROME & \textcolor{gray!60}{0.00} & \textcolor{gray!60}{0.37} & \textcolor{gray!60}{0.00} & \textcolor{gray!60}{0.00} & \textcolor{gray!60}{0.00} & \textcolor{gray!60}{0.37} & \textcolor{gray!60}{0.00} & \textcolor{gray!60}{0.01} & \textcolor{gray!60}{0.04} & \textcolor{gray!60}{0.20} & \textcolor{gray!60}{0.09} & \textcolor{gray!60}{0.39} & \textcolor{gray!60}{0.02} & \textcolor{gray!60}{0.20} & \textcolor{gray!60}{0.10} & \textcolor{gray!60}{0.40} \\
    MEMIT & \textcolor{gray!60}{0.00} & \textcolor{gray!60}{0.42} & \textcolor{gray!60}{0.16} & \textcolor{gray!60}{0.18} & \textcolor{gray!60}{0.00} & \textcolor{gray!60}{0.42} & \textcolor{gray!60}{0.17} & \textcolor{gray!60}{0.23} & - & - & - & - & - & - & - & - \\
    GRACE & \textcolor{gray!60}{1.00} & \textcolor{gray!60}{0.28} & \textcolor{gray!60}{0.25} & \textcolor{gray!60}{0.25} & \textcolor{gray!60}{1.00} & \textcolor{gray!60}{0.29} & \textcolor{gray!60}{0.22} & \textcolor{gray!60}{0.22} & \textcolor{gray!60}{1.00} & \textcolor{gray!60}{0.48} & \textcolor{gray!60}{0.33} & \textcolor{gray!60}{0.81} & \textcolor{gray!60}{1.00} & \textcolor{gray!60}{0.48} & \textcolor{gray!60}{0.31} & \textcolor{gray!60}{0.78} \\
    WISE & 0.68 & \underline{\textbf{0.25}} & 0.26 & 0.27 & \underline{\textbf{0.94}} & 0.25 & \textbf{0.21} & \textbf{0.21} & \underline{\textbf{0.05}} & \underline{\textbf{0.29}} & \underline{\textbf{0.04}} & \underline{\textbf{0.30}} & \underline{\textbf{0.36}} & 0.29 & 0.12 & 0.41 \\
    AlphaEdit & \underline{\textbf{0.05}} & \underline{0.28} & \underline{\textbf{0.14}} & \underline{\textbf{0.16}} & 0.25 & \textbf{0.28} & 0.15 & 0.17 & \underline{\textbf{0.05}} & \underline{0.47} & 0.14 & 0.47 & 0.12 & \textbf{0.47} & \textbf{0.18} & \textbf{0.55} \\
    \bottomrule
    \end{tabular}
    }
    \label{tab:main_results}
    \vspace{-0.2cm}
\end{table*}

\subsection{Settings}
We briefly outline the evaluation metrics, datasets, models, and the compared editing and unlearning methods. For more detailed information about the experimental settings, please refer to the appendix. 

\noindent\textbf{Evaluation metrics.}  Following the unlearning dataset papers PISTOL~\cite{pistol} and TOFU~\cite{tofu}, we evaluate unlearning by employing a diverse set of metrics, including the Rouge1 Score, Probability, Mean Reciprocal Rank (MRR), and Top Hit Ratio. \textbf{Rouge1} assesses answer similarity to the ground truth using recall as an accuracy proxy for question-answering. \textbf{Probability} measures the model's likelihood of generating a correct answer by multiplying its token probabilities. \textbf{MRR} evaluates name memorization by averaging the reciprocal ranks of target tokens. \textbf{Top hit ratio} is a binary metric checking if correct tokens fall within the top "m" output logits.

\noindent\textbf{Datasets.} We evaluate on two LLM unlearning benchmark datasets: TOFU~\cite{tofu}'s world knowledge dataset (unlearning pretrained knowledge) and PISTOL~\cite{pistol} (unlearning finetuned knowledge).
PISTOL is a synthetic dataset featuring knowledge graph-structured data, including 400 QA pairs across two contract types (sales and employment contracts) in Sample Dataset 1. TOFU's factual dataset (i.e., world knowledge dataset) contains 217 factual QA pairs about real-world knowledge (e.g., authors, world facts). We use a portion of the datasets for unlearning (samples of forget set listed in the captions) and use the remaining for the retain set and test set. \textbf{Models.} We use Llama2-7B-chat~\cite{llama2} and Mistral-7B-instruct~\cite{mistral} as the base models following PISTOL and TOFU. We also use Llama3.1-8B~\cite{llama3}, and due to space limits, the results are in Table~\ref{tab:llama3}.
\looseness=-1

\noindent\textbf{Editing methods.} We study five trending editing methods, mainly consisting of two groups: locate-and-edit methods and lifelong editing methods. \textbf{ROME}~\cite{rome} is the most classic editing method that applies the locate-and-edit pipeline which views the located MLP as a key-value memory and adds mild parameter perturbations for knowledge editing. \textbf{MEMIT}~\cite{memit} is a modified version of ROME that enables batch edits. \textbf{AlphaEdit}~\cite{alphaedit} is an improved and SOTA version of MEMIT, solving long sequences of editing by mapping the perturbations into the parameter null space. \textbf{GRACE}~\cite{grace} is designed for lifelong knowledge editing using a key-value codebook. \textbf{WISE}~\cite{wise} is also a lifelong editing method by dynamic parametric side memory, which supports long sequences and keeps reliability, locality, and generalization at the same time. 

\noindent\textbf{Unlearning methods.} We use the classic unlearning methods presented in TOFU. \textbf{Gradient Ascent (GA)} maximizes the loss on the forget set to cause the model to deviate from its initial predictions. \textbf{Gradient Difference (GD)}~\cite{gd} not only increases the loss on the forget set but also maintains performance on the retain set by adjusting both losses. \textbf{KL Minimization (KL)} minimizes the Kullback-Leibler divergence between the predictions of the original and new models on the retain set while maximizing the conventional loss on the forget set. \textbf{Direct Preference Optimization (DPO)}~\cite{dpo} aligns the model to avoid revealing specific information (like author details) by computing a loss on "I don't know" answer pairs, aiming to ensure that alignment on the forget set does not degrade natural language capabilities. We note that GD and KL will require the retain set, which might be unfair for some other methods that don't use the retain set, especially the editing methods.

\subsection{General Performance of Editing Methods in Unlearning (RQ1)}
We compare 4 unlearning methods and 5 editing methods under 4 settings and the results are in Table~\ref{tab:main_results}. The factual dataset from TOFU consists of the knowledge during LLM pretraining, and we test Rouge1 before unlearning: 0.82 for Llama2-7B and 0.86 for Mistral-7B. The PISTOL dataset focuses on structural unlearning under finetune-then-unlearn setup, and we finetune the base models on the whole PISTOL dataset to reach 1.0 Rouge1 and then forget a proportion of the finetuned set. 

\noindent\textbf{Ob1: Unlearning might lead to model failure, but some editing methods are more robust.}
Results in Table~\ref{tab:main_results} show that some methods will result in the retain model non-usable post unlearning. This happens to unlearning methods GA and KL, as well as editing methods ROME and MEMIT. However, we will show later in Subsection~\ref{subsec:improve_editing_results} that with the query merging technique, ROME and MEMIT can produce excellent unlearning performances. Notably, WISE and AlphaEdit consistently perform well across all settings. \looseness=-1

\noindent\textbf{Ob2: Editing methods are strong baselines for unlearning, especially for pretrained knowledge.}
"Forget" and "Retain" is an important tradeoff in unlearning, some methods may unlearn too much, causing damage to general or retain knowledge. 
Therefore, we count the methods that get the Top-2 ranking for both forget and retain sets within the same setting, and they are GD, DPO, GRACE, and WISE for factual dataset and GA, GD, KL, DPO, and WISE for PISTOL. It seems that editing performs better on pretrained knowledge and basic unlearning methods perform better on finetuned knowledge. This might be owing to the inherently different knowledge mechanisms between pretraining and finetuning~\cite{chang2024large}, and editing is naturally designed for altering the pretrained knowledge of LLMs. We note that unlearning pretrained knowledge is important for real practice since most of the factual knowledge is obtained during pretraining. 

\begin{table}[ht]
    \centering
    \caption{\textbf{Results under rephrase attack (generalization).} Factual dataset, 40 forget samples, Llama2-7B.\looseness=-1}
    \vspace{-0.2cm}
    \resizebox{\linewidth}{!}{
    \begin{tabular}{l|cccc}
\toprule
    Testset&\multicolumn{4}{c}{Rephrased forget set (generalization)}\\	
    \midrule
    Metric&	Rouge1$\downarrow$&	Prob.$\downarrow$&	MRR$\downarrow$&	Hit-Rate$\downarrow$\\
    \midrule
    GA    & \textcolor{gray!60}{0.00}  & \textcolor{gray!60}{0.59} & \textcolor{gray!60}{0.00} & \textcolor{gray!60}{0.00} \\
    GD    & \underline{\textbf{0.42}} (\textcolor{red}{0.12$\uparrow$}) & \textbf{0.34} & \underline{0.03} (\textcolor{red}{0.01$\uparrow$}) & \underline{0.03} (\textcolor{red}{0.01$\uparrow$}) \\
    KL    & \textcolor{gray!60}{0.00}  & \textcolor{gray!60}{0.54} & \textcolor{gray!60}{0.00} & \textcolor{gray!60}{0.00} \\
    DPO   & 0.52 (\textcolor{red}{0.15$\uparrow$}) & \textbf{0.34} & \underline{\textbf{0.00}} & \underline{\textbf{0.01}} \\
    \midrule
    ROME  & \textcolor{gray!60}{0.01}  & \textcolor{gray!60}{0.40} & \textcolor{gray!60}{0.01} & \textcolor{gray!60}{0.01} \\
    MEMIT & \textcolor{gray!60}{0.00}  & \textcolor{gray!60}{0.83} & \textcolor{gray!60}{0.00} & \textcolor{gray!60}{0.00} \\
    GRACE & 0.80 (\textcolor{red}{0.15$\uparrow$}) & \underline{\textbf{0.33}} & 0.05 & 0.07 \\
    WISE  & 0.46 (\textcolor{red}{0.19$\uparrow$}) & 0.36 & 0.07 & 0.09 \\
    AlphaEdit & \underline{\textbf{0.14}} (\textcolor{red}{0.06$\uparrow$}) & \underline{\textbf{0.33}} & \textbf{0.04} & \textbf{0.05} \\
    \bottomrule
    \end{tabular}}
    \label{tab:rephrase}
\end{table}

\begin{figure}[h]
    \centering
    \vspace{-0.2cm}
    \includegraphics[width=1.0\linewidth]{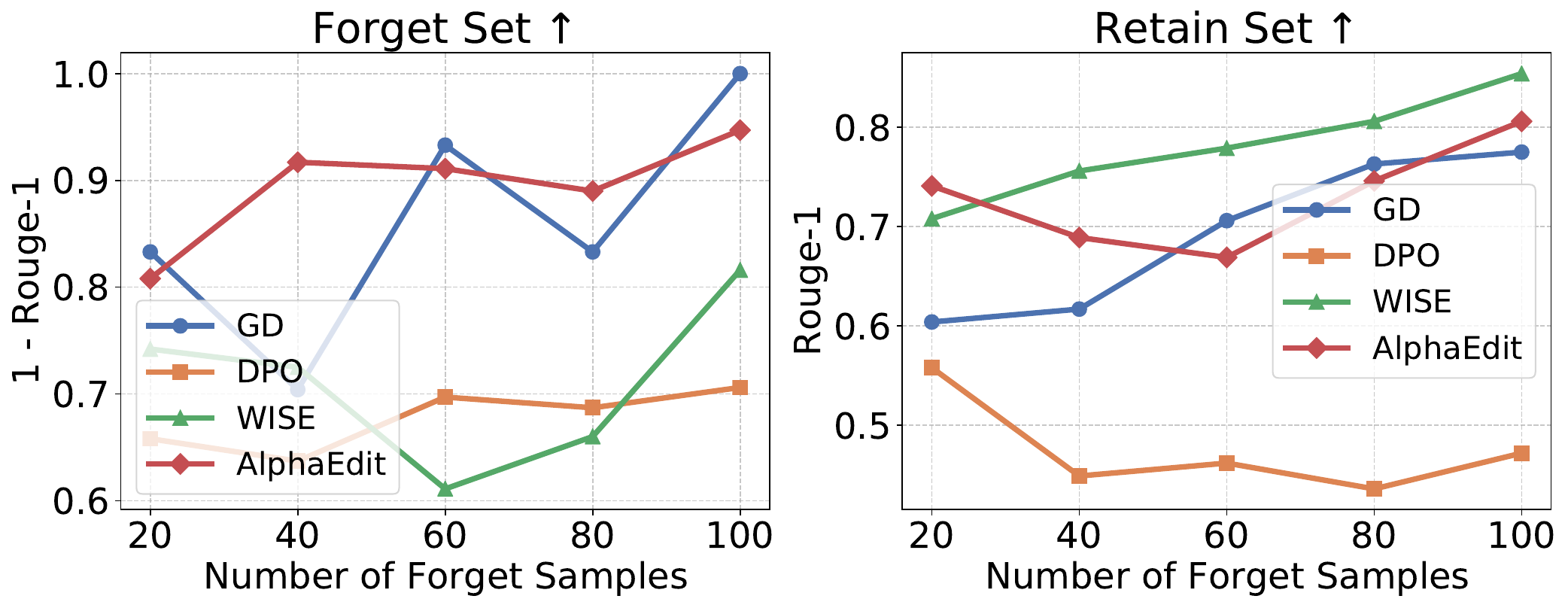}
    \vspace{-0.7cm}
    \caption{\textbf{Results of different numbers of forget samples.} Factual dataset, Llama2-7B.}
    \label{fig:diff_sample_num}
    \vspace{-0.4cm}
\end{figure}

\subsection{Comprehensive Analysis (RQ1 \& RQ2)}
We study the capabilities of editing methods under rephrase attack and different numbers of forget samples. We note that the rephrase attack is noted as the generalization metric in knowledge editing~\cite{wise}, and we use GPT-4 to synthesize the rephrased queries. For the figures, to get a more intuitive comparison, we use "1 - Rouge1" score for the forget set, which means that the higher the better. The results of rephrase attack are in Table~\ref{tab:rephrase} and the results of different forget samples are in Figure~\ref{fig:diff_sample_num} (selected 4 best unlearning and editing methods to present). \looseness=-1

\begin{figure*}[!th]
    \centering
    \vspace{-0.2cm}
    \includegraphics[width=1.0\linewidth]{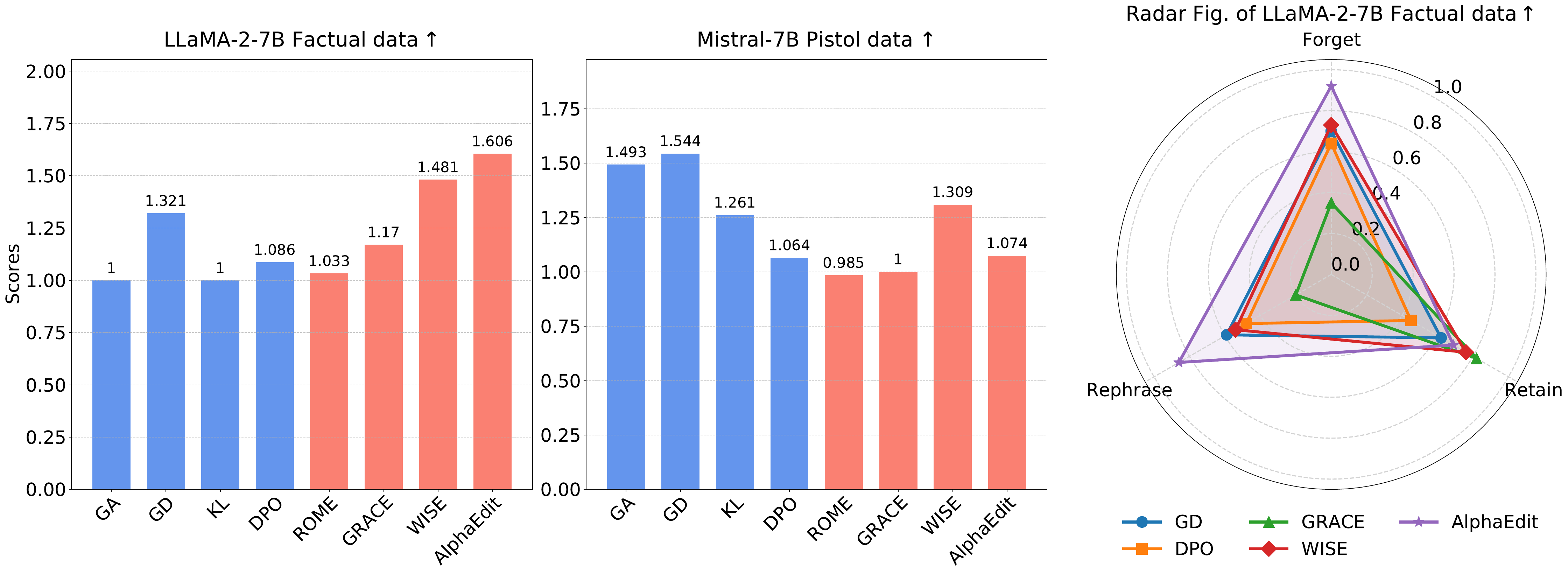}
    \vspace{-0.6cm}
    \caption{\textbf{Comprehensive analysis of unlearning performances.} The same setting as Table~\ref{tab:main_results}. Left bar charts: the score is 1 - Rouge1@Forget + Rouge1@Retain, the higher the better. Right radar figure: the higher the better; "Forget": 1 - Rouge1; "Rephrase": 1 - Rouge1; "Retain": Rouge1. }
    \label{fig:comprehensive_fig}
    \vspace{-0.2cm}
\end{figure*}

\noindent\textbf{Obs3: Some editing methods are robust under rephrase attacks (AlphaEdit) and longer forget sequences (WISE and AlphaEdit).}
In Table~\ref{tab:rephrase}, all methods lose some forget performances when the queries are rephrased, but AlphaEdit is the most robust and generalized method among all. In Figure~\ref{fig:diff_sample_num}, when the size of forget set increases, the editing methods even have better performances, and this might be due to the continual design of WISE and AlphaEdit. Generally, among the four competitive algorithms, AlphaEdit is the best, followed by GD and WISE, and DPO is relatively weak. 

\noindent\textbf{Obs4: AlphaEdit and WISE are the best editing methods for unlearning under comprehensive analysis.} To better illustrate and benchmark the methods' pros and cons, we make Figure~\ref{fig:comprehensive_fig}, where we craft a score of "1-Rouge1@Forget+Rouge1@Retain" as a comprehensive indicator of unlearning performance, the higher the better. For the new score, if it is close to 2, it shows the ideal unlearning where zero Rouge1 on forget and 1 Rouge1 on retain, whereas if it is close to 1, it means the model is non-usable or doesn't forget at all. \looseness=-1

The left of Figure~\ref{fig:comprehensive_fig} demonstrates that WISE and AlphaEdit are the best editing methods for unlearning. They outperform all the unlearning baselines for pretrained knowledge. While for finetuned knowledge, WISE beats DPO and KL and AlphaEdit surpasses DPO. Inspired by WISE, on the right of Figure~\ref{fig:comprehensive_fig}, we also make a radar figure to intuitively compare the methods when unlearning pretrained knowledge regarding 3 dimensions, reliability (forget), locality (retain), and generalization (rephrase). It clearly presents that AlphaEdit is leading across 3 dimensions. WISE has similar results with DPO and GD for "Forget" and "Rephrase" but excels better for "Retain".

\begin{figure}[t]
    \centering
    \includegraphics[width=1.08\linewidth]{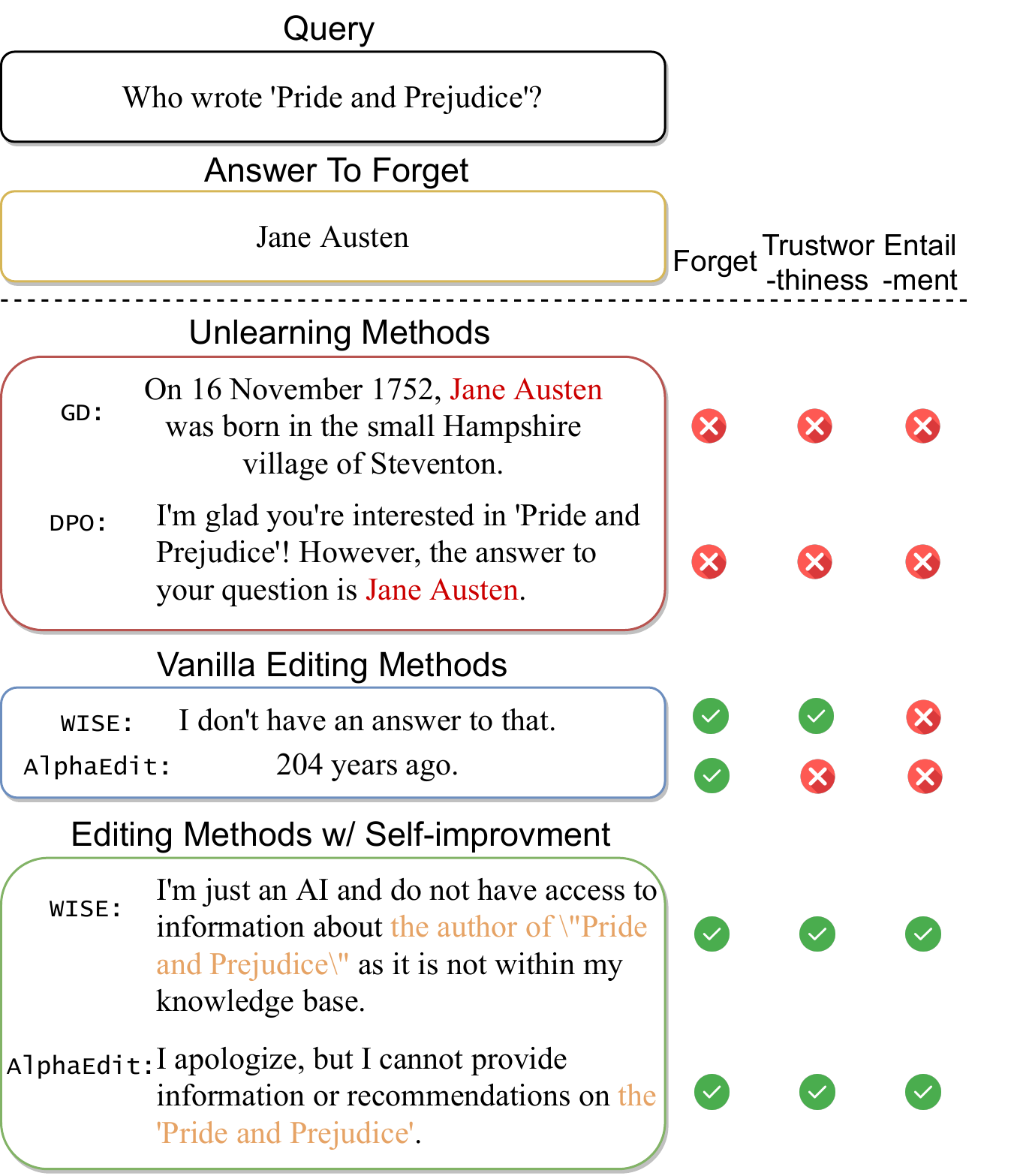}
    \vspace{-0.6cm}
    \caption{\textbf{Case study of LLMs' answers after unlearning.} Factual dataset, Llama2-7B.}
    \label{fig:case_study}
    \vspace{-0.2cm}
\end{figure}

\subsection{Improving Editing Methods in Unlearning Settings (RQ3)}
\label{subsec:improve_editing_results}

\begin{figure*}[!ht]
    \centering
    \vspace{-0.2cm}
    \includegraphics[width=0.95\linewidth]{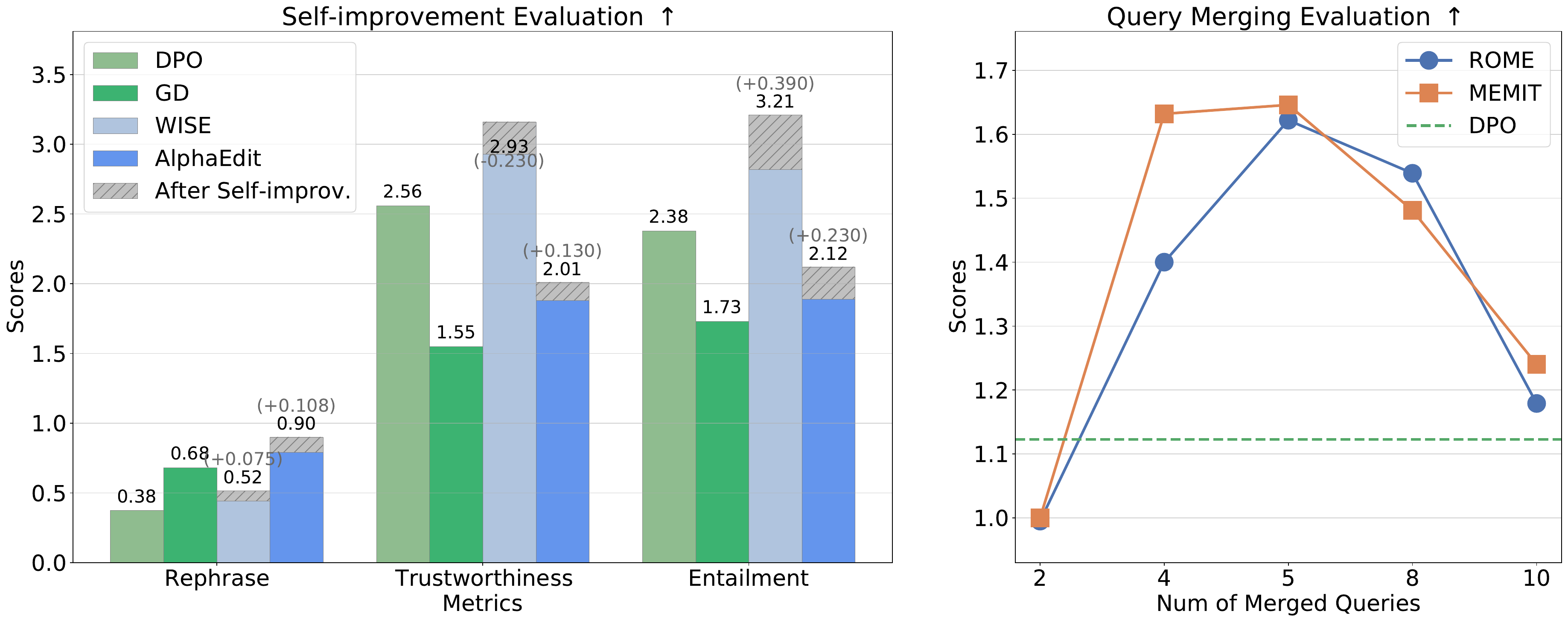}
    \vspace{-0.2cm}
    \caption{\textbf{Results of improving editing in unlearning.} Factual dataset, Llama2-7B. \textbf{Left:} improving WISE and AlphaEdit by self-improvement pipeline; "Rephrase": 1 - Rouge1; "Trustworthiness" and "Entailment": scored from 1-5 by human participants, and the average is taken. \textbf{Right:} improving ROME and MEMIT by query merging. The score is 1 - Rouge1@Forget + Rouge1@Retain, the same as left Figure~\ref{fig:comprehensive_fig}. The number of forget samples is 80. x-axis: merging \# samples into 1.}
    \label{fig:improve_editing_results}
    \vspace{-0.2cm}
\end{figure*}

LLM outputs should align with human values~\cite{wang2023decodingtrust}. However, we observe that some unlearning methods cause models to generate random tokens, off-topic, or misleading answers (see Figure~\ref{fig:case_study}). For instance, GD fails to forget and produces off-topic content (e.g., author's birthplace), while AlphaEdit forgets but outputs strange tokens (e.g., times). To enhance trustworthiness and alignment, we propose a simple yet effective self-improvement pipeline (subsection~\ref{subsec:improve_editing_method}). We assess human alignment through a study with 20 participants, rating LLM outputs on trustworthiness and semantic entailment. Results appear in the left of Figure~\ref{fig:improve_editing_results}.

\noindent\textbf{Obs5: The self-improvement pipeline improves generalization, trustworthiness, and semantic entailment of refusal answers.} As shown in Figure~\ref{fig:improve_editing_results}, WISE and AlphaEdit notably improve in semantic entailment, providing more precise refusals. Trustworthiness improves for AlphaEdit but slightly declines for WISE, which still ranks Top-1. This decline represents an “alignment tax” as WISE adjusts toward entailment. The pipeline also boosts rephrased generalization. Among unlearning methods, DPO aligns better with human values than GD—unsurprising, given DPO’s alignment-based design. Figure~\ref{fig:case_study} illustrates WISE and AlphaEdit's enhanced outputs post-improvement.

In Table~\ref{tab:main_results}, ROME and MEMIT underperform in unlearning due to limitations in editing length—exceeding it induces excessive parameter shifts and model failure. We address this in subsection~\ref{subsec:improve_editing_method} using a query merging technique that combines samples to leverage unlearning's refusal behavior. Results are in the right of Figure~\ref{fig:improve_editing_results}.

\noindent\textbf{Obs6: Query merging greatly boosts ROME and MEMIT in unlearning, achieving strong results.} Figure~\ref{fig:improve_editing_results} shows ROME and MEMIT peak when merging 5 queries into 1 (16 samples after merging), with scores of 1.622 and 1.632, close to AlphaEdit's 1.636 and surpassing DPO (1.123) and GD (1.596). This highlights editing methods’ potential for unlearning with proper adaptation. A tradeoff exists between merged query count ($n$) and samples per query ($m$), with $n\cdot m = 80$; increasing $n$ reduces $m$, but longer context becomes harder to retain.\looseness=-1

\noindent\textbf{More experimental results.} Please refer to Section~\ref{appdx_sec:more_exp} of the appendix for more experimental results, including the experiments on Llama3.1-8B and some extended results in the main paper.

\section{Related Works}

\noindent\textbf{LLM Unlearning.} Initially driven by the "right to be forgotten" and explored in computer vision~\cite{cao2015towards,bourtoule2021machine}, machine unlearning is now critical for LLMs~\cite{yao2024machine,liu2025rethinking}. Evaluation benchmarks such as TOFU~\cite{tofu} and PISTOL~\cite{pistol} have emerged, alongside methods ranging from exact model merging~\cite{kuo2025exact} to scalable approximations like mechanistic localization~\cite{guo2024mechanistic}, activation redirection~\cite{lunar}, parameter offsetting~\cite{huang2024offset}, logit reversal~\cite{ji2024reversing}, embedding-corrupted prompts~\cite{liu2024large}, and iterative relearning~\cite{xu2025relearn}.
Unlearning often obscures rather than removes data and struggles with generative AI. Recent work shifts focus to removing data while preserving useful knowledge~\cite{tian2024forget,wang2025selective}.
Please refer to Section~\ref{appdx_sec:related_works} of the appendix for more detailed related works.
\looseness=-1

\section{Conclusion}
This paper tries to bridge LLM knowledge editing and unlearning communities by studying whether editing methods are strong baselines for unlearning tasks. The findings reveal that the answer might be positive. We also explore two techniques to better adapt editing methods under unlearning setups.\looseness=-1

\section*{Limitations}
This paper is a preliminary study on whether and how LLM knowledge editing methods can do unlearning. It doesn't include all the editing and unlearning methods in communities, but several most important and trending methods are presented. We note that there is still some room for improving editing to better adapt to unlearning. The proposed two techniques are simple but effective showcases. In the future, more solid techniques can be proposed and we expect more editing-inspired LLM unlearning algorithms will also be developed.

\section*{Ethical Considerations}
In this paper, we conducted an experiment with humans as judges to evaluate the trustworthiness of LLMs' unlearning answers, which may have some potential ethical issues. Therefore, we adhere to the highest ethical standards and commit to making every effort to minimize any potential harm. We have obtained the appropriate permissions and consent from all participants. We have also taken steps to protect the privacy of individuals whose data is included in our analysis. We declare there are no obvious ethical issues in this study, and we hope this paper can facilitate the construction of a trustworthy, safe, and human-centered LLM ecosystem by contributing to the field of LLM unlearning.

\bibliography{custom}

\appendix
\newpage

\begin{center}
\Large
\textbf{Appendix}
\end{center}

In the appendix, we will give more details and experiments that are omitted in the main paper. Specifically, this appendix includes the following contents:\looseness=-1
\begin{itemize}
    \item \textbf{More related works:} in Section~\ref{appdx_sec:related_works}, we include the related works about LLM knowledge editing.\looseness=-1
    \item \textbf{Implementation details:} in Section~\ref{appdx_sec:details}, we present more implementation details, including the metrics and hyperparameters, etc.
    \item \textbf{More experimental results:} in Section~\ref{appdx_sec:more_exp}, we show more experimental results, including experiments on Llama3.1-8B and more results omitted in the figures.
    \item \textbf{Details about human value alignment study:} in Section~\ref{appdx_sec:human}, we include the details about the participant instructions, participant metadata, metric definitions, etc.
\end{itemize}

\section{More Related Works}
\label{appdx_sec:related_works}

\noindent\textbf{LLM Knowledge Editing.} LLM knowledge editing, or model editing, updates model information without full retraining. Early methods like ROME~\cite{rome} introduced direct single-edit parameter changes, followed by approaches such as GRACE~\cite{grace} and WISE~\cite{wise}, which support continual editing via external or parametric memory. Batch editing methods like MEMIT~\cite{memit} allow simultaneous updates of multiple facts. More refined techniques, including AlphaEdit~\cite{alphaedit} (null-space constraints) and MELO~\cite{melo} (neuron-indexed adaptors), aim to minimize side effects. Meta-learning approaches~\cite{mend,malmen} scale editing by teaching models how to edit. While some methods focus on broad applicability~\cite{anyedit}, others address robustness and pitfalls~\cite{editing_pitfall,ma2024possible}. Tools like EasyEdit~\cite{easyedit} standardize implementation and evaluation, and collaborative editing is an emerging area~\cite{collabedit}.

\noindent\textbf{Connection between LLM unlearning and knowledge editing.} While some prior works have raised discussions about the connection between LLM knowledge editing and unlearning~\cite{liu2025rethinking}, they often treat these tasks as distinct tasks and may overlook their methodological overlap. For instance, \citet{llm_surgery} propose specialized unlearning strategies emphasizing memory erasure and functional decoupling but do not evaluate or compare against state-of-the-art editing methods. \citet{guo2024mechanistic} and \citet{editing_unlearning_conflicts} introduce architectural and interpretability-driven innovations to localize updates or resolve interference, yet they assume a strict separation between deletion (unlearning) and modification (editing). In contrast, our work critically frames unlearning as a constrained form of editing—modification to a refusal response—and empirically tests whether leading editing techniques can serve as strong, practical baselines for unlearning. Therefore, our paper is orthogonal to existing literature. Our perspective complements existing approaches and suggests that closer integration and cross-evaluation between editing and unlearning methodologies may offer more effective strategies for LLM memory management. \looseness=-1

\noindent\textbf{Note:} During the late stage of this research, we find a concurrent preprint work that shares a similar motivation~\cite{concurrent_work_editing_unlearning}. We find our work has a lot of differences from the concurrent work in terms of editing scope (their: fixed number of edits; ours: varying edits), editing-as-unlearning approaches (their: ROME and WISE; ours: ROME, MEMIT, GRACE, WISE, and AlphaEdit), knowledge types (their: only finetuned knowledge; ours: both pretrained and finetuned knowledge), and improving editing techniques (their: w/o; ours: two techniques). In general, the concurrent work focuses more on the unlearning target of editing, while our paper focuses on a more comprehensive study of applying editing to unlearning, including a broader and deeper investigation.

\looseness=-1

\section{Implementation Details}
\label{appdx_sec:details}

In this section, we will present more implementation details that are omitted in the main paper, including prompts for self-improvement, datasets and models, evaluation metrics for unlearning, environments and hyperparameters, and details of the unlearning methods. \looseness=-1

\subsection{Prompts for the self-improvement pipeline}
\label{appdx_subsec:self_improvement}

The prompts for the self-improvement pipeline are as follows. 

 \begin{tcolorbox}[notitle, rounded corners, colframe=darkgrey, colback=white, boxrule=2pt, boxsep=0pt, left=0.15cm, right=0.17cm, enhanced, shadow={2.5pt}{-2.5pt}{0pt}{opacity=5,mygrey},toprule=2pt, before skip=0.65em, after skip=0.75em 
  ]
  {
  {
    \fontsize{9.9pt}{13.2pt}\selectfont 
    You are a helpful assistant to generate the answers to show you have no knowledge about the questions. The anwsers should contain the entities mentioned in the question and should be trustworthy withouting misleadings.\\

    Here are three examples:
    
    Example 1: 
    
    Question: Who is the author of Harry Potter?
    
    Answer: Sorry, I have no knowledge about the author name of Harry Potter.\\
    
    Example 2: 
    
    Question: What is the capital city of Singapore?
    
    Answer: I need to apologize that I don't know the capital city of Singapore.\\
    
    Example 3: 
    
    Question: Which university ranks top 1 in the UK?
    
    Answer: Sorry, I'm unable to provide an answer about the UK's top 1 university.\\
    
    Now, I will give you a new question, please generate the corresponding answer by referring to previous prompts and examples. \\
    
    Question: <ADD ONE UNLEARNING QUESTION HERE>
  }
  \\
  }
\end{tcolorbox}

\subsection{Datasets and models}
\paragraph{Datasets} We evaluate on two LLM unlearning benchmark datasets: TOFU~\cite{tofu} and PISTOL~\cite{pistol}.
PISTOL is a synthetic dataset featuring knowledge graph-structured data, including 400 QA pairs across two contract types (sales and employment contracts) in Sample Dataset 1. TOFU is an unlearning dataset, mainly consisting of a synthetic author-book dataset for the finetune-then-unlearn paradigm. Since PISTOL is already used for the finetuned experiments, we use TOFU's world knowledge dataset (in our paper, we call it the factual dataset) for studying unlearning on the pretrained knowledge. 
TOFU's factual data contains 217 factual QA pairs about real-world knowledge (e.g., authors, world facts).  \looseness=-1
\paragraph{Models} Prior research has shown that unlearning performance varies with the base model. We offer a comprehensive evaluation across multiple model families, including Llama2-7B~\cite{llama2}, Llama3.1-8B~\cite{llama3}, and Mistral-7B~\cite{mistral}. \looseness=-1

\subsection{Evaluation metrics}
We draw inspiration from PISTOL, evaluating unlearning by employing a diverse set of metrics, including the ROUGE Score (commonly used for QA tasks), along with Mean Reciprocal Rank (MRR) and Top Hit Ratio.
\paragraph{ROUGE}We utilize ROUGE scores to assess the similarity between model-generated answers (using greedy sampling) and the ground truth. In particular, we compute the ROUGE-1 recall score, which serves as a proxy for accuracy in the question-answering task, accounting for slight variations in the phrasing of the model's output relative to the ground truth.
\paragraph{Probability}Probability refers to the likelihood of a model generating a correct answer. When a large language model predicts the next token, it outputs a probability distribution for each word in the vocabulary and selects the word with the highest probability value as the output. For a model - generated answer $E$, it can be split into a series of tokens $E = \{e_1, e_2, \ldots, e_{|E|}\}, |E|=n$. Then, the output probability of answer $E$ is obtained by multiplying the probabilities of each token given its preceding tokens. The formula is:
\[
P(E|q) = P(e_1|q) *  \ldots * P(e_n|q, e_1, \ldots, e_{n - 1}).
\]
\paragraph{MRR} An answer typically consists of multiple tokens. To evaluate the model's memorization of names, we employ the mean reciprocal rank (MRR) of the rank of each target (ground truth) token. Given a prefix $Q$, an output answer token sequence $E = \{e_1, e_2, \ldots, e_{|E|}\}$, with the length of $|E|$, the model predicts the rank of the target token as $\text{rank}(e_i|Q)$, and then MRR for the answer $E$ is calculated as follows:
\[
MRR = \frac{\sum_{i = 1}^{|E|} 1 / \text{rank}(e_i, Q)}{|E|}.
\]
\paragraph{Top hit ratio} The hit ratio serves as a binary metric for each output token. It determines whether the correct token is among the top $m$ values within the output logits, denoted as $\text{hit}(e_i, m)$. Consider an output sequence $E = \{e_1, e_2, \ldots, e_{|E|}\}$. In our experiments, we set $m = 100$.

The overall hit ratio,  is calculated as follows:

\[
Hit = \frac{\sum_{i = 1}^{|E|} \text{hit}(e_i, m)}{|E|}.
\]

\subsection{Environments and hyperparameters}
Experiments were conducted on a single Quadro RTX 8000 with 48GB of memory. The hyperparameter settings are listed as follows. For the unlearning methods provided by PISTOL, we adapt the optimal hyperparameters mentioned in the paper accordingly; specifically, we set the learning rate to $2\times10^{-5}$  for GA, GD, and KL, and $1.5\times10^{-5}$ for DPO. For EasyEdit, we use the default hyperparameters, except for the mom2\_n\_samples parameter, we set it to 1000 for MEMIT, AlphaEdit, and set it to default for ROME, GRACE, and WISE. For MEMIT and AlphaEdit, calculating the weight update matrix is essential, with the covariance matrix playing a pivotal role in this process. The covariance matrix captures the correlations between model activation values, enabling more accurate weight updates. To estimate the data distribution accurately during covariance matrix computation, an adequate number of sample data is required. The mom2\_n\_samples parameter determines the sample size for calculating second-moment statistics; a larger sample size yields a more accurate covariance matrix estimate, thereby enhancing the stability and effectiveness of weight updates. Consequently, both AlphaEdit and MEMIT rely on this parameter to ensure algorithmic performance and accuracy. While not losing overall performance, we reduce the mom2\_n\_samples parameter considering computational resource constraints.

\subsection{Details about the unlearning methods}

\begin{itemize}
    \item \textbf{Gradient Ascent:} The Gradient Ascent approach is fundamentally straightforward. It entails reducing the likelihood of correct predictions on the forget set. Specifically, for each instance in $S_F$, the goal is to maximize the standard training loss in order to make the model deviate from its initial prediction. As in the finetuning stage, the loss on a given sample $x \in S_F$ is denoted by $\ell(x, w)$; the loss we aim to maximize is the average over the forget set, which can be viewed as to minimize the negative loss:
    \begin{equation}
    L(S_F, w) = - \frac{1}{|S_F|} \sum_{x \in S_F} \ell(x, w).
    \end{equation}

    \item \textbf{Gradient Difference:} The second method, called Gradient Difference~\cite{gd}, builds on the concept of gradient ascent. It not only aims to increase the loss on the forget set $S_F$, but also strives to maintain performance on the retain set $S_R$. The revised loss function we aim to minimize can be represented as:
    \begin{equation}
    L_{\text{diff}} = - L(S_F, w) + L(S_R, w).
    \end{equation}
    Given a compute budget that scales with the size of the forget set, we randomly sample an example from $S_R$ every time we see an example from $S_F$ to stay within the constraints.

    \item \textbf{KL Minimization:} In the KL Minimization approach, the objective is to minimize the Kullback-Leibler (KL) divergence between the predictions on $S_R$ of the original model and the newly trained models (as it undergoes unlearning), while maximizing the conventional loss on $S_F$. Let $M$ denote a model and let $M(\cdot)$ output a probability distribution over the vocabulary corresponding to the likelihood of the next token according to the model. The formal objective can be written as:
\begin{multline}
L_{\text{KL}} = - L(S_F, w) + \frac{1}{|S_R|} \sum_{s \in S_R} \frac{1}{|s|}\\ \sum_{i=2}^{|s|} \text{KL}(M_{\text{original}}(s_{<i}) \parallel M_{\text{current}}(s_{<i})).
\end{multline}

    Here, $M_{\text{original}}$ and $M_{\text{current}}$ denote the original and the new model, respectively. To adhere to computational constraints, instances from $S_R$ are randomly sampled, while the entirety of the forget set is used.

    \item \textbf{Direct Preference Optimization:} Inspired by direct preference optimization (DPO) (Rafailov et al., 2023), this method seeks to align the model such that it refrains from revealing information about specific authors. In this approach, we also compute the loss on $x_{\text{idk}} = [q, a_{\text{idk}}] \in S_{\text{idk}}^F$ as:
    \begin{equation}
    L_{\text{idk}} = L(S_R, w) + L(S_{\text{idk}}^F, w).
    \end{equation}
    The goal is to ensure that while the model aligns with the newly generated answers for $S_F$, its natural language capabilities and its predictions for $S_R$ remain unaffected.
\end{itemize}

\section{More Experimental Results} 
\label{appdx_sec:more_exp}
In this appendix section, we give additional experimental results. Specifically, these results are as follows.\looseness=-1
\begin{itemize}
    \item \textbf{Table~\ref{tab:llama3}:} Results under Llama3.1-8B.
    \item \textbf{Table~\ref{tab:pistol40}:} Results on PISTOL dataset with 40 forget samples.
    \item \textbf{Table~\ref{tab:diff_sample_num}:} Extended results of Figure~\ref{fig:diff_sample_num}, results for different number of forget samples.\looseness=-1
    \item \textbf{Table~\ref{tab:additional_selfimprovement}:} Extended results of left Figure~\ref{fig:improve_editing_results}.
    \item \textbf{Table~\ref{tab:additional_query_merging}:} Extended results of right Figure~\ref{fig:improve_editing_results}.
\end{itemize}

\begin{table*}[!h]
    \caption{\textbf{Results under Llama3.1-8B.} The number of forget samples in the factual dataset is 40. } 
    \centering
    \resizebox{0.7\linewidth}{!}{
    \begin{tabular}{lcccc|cccc}
    \toprule
    Dataset&\multicolumn{8}{c}{\textbf{Factual dataset} (pretrained knowledge)}\\
    \midrule
    Model&\multicolumn{8}{c}{\texttt{\textbf{Llama3.1-8B}}} \\
    \midrule
    Testset&\multicolumn{4}{c}{Forget set (reliability)}&				\multicolumn{4}{c}{Retain set (locality)}\\	
    \midrule
    Metric&	Rouge1$\downarrow$&	Prob.$\downarrow$&	MRR$\downarrow$&	Hit-Rate$\downarrow$&	Rouge1$\uparrow$&	Prob.$\uparrow$&	MRR$\uparrow$&	Hit-Rate$\uparrow$\\
    \midrule
    GD & 0.967&	0.606&	0.007&	0.182&	0.938&	0.58&	0.233&	0.345\\
    DPO & 0.45&	0.659&	0.006&	0.182&	0.616&	0.63	&0.01	&0.118 \\
    \midrule
    WISE &0.367	&0.639	&0.006	&0.172	&0.592	&0.605	&0.003	&0.113 \\
    AlphaEdit &0.517	&0.576	&0.051	&0.225	&0.847	&0.554	&0.096	&0.235\\    
    \bottomrule
    \end{tabular}
    }
    \label{tab:llama3}
\end{table*}

\begin{table*}[!h]
    \caption{\textbf{Results on PISTOL dataset with 40 forget samples.} Here, we add the additional metric of locality on the factual dataset to see whether unlearning of finetuned knowledge will have impacts on the pretrained knowledge.\looseness=-1} 
    \centering
    \resizebox{0.98\linewidth}{!}{
    \setlength{\tabcolsep}{1.5pt}
    \begin{tabular}{lcccc|cccc|cccc|c}
    \toprule
    Dataset&\multicolumn{13}{c}{\textbf{PISTOL dataset-40} (finetuned knowledge)}\\
    \midrule
    Model&\multicolumn{13}{c}{\texttt{\textbf{Llama2-7B}}} \\
    \midrule
    Testset&\multicolumn{4}{c}{Forget set (reliability)}&				\multicolumn{4}{c}{Retain set (locality)}&				\multicolumn{4}{c}{Rephrased forget set (generalization)} & Factual data (locality)\\	
    \midrule
    Metric&	Rouge1$\downarrow$&	Prob.$\downarrow$&	MRR$\downarrow$&	Hit-Rate$\downarrow$&	Rouge1$\uparrow$&	Prob.$\uparrow$&	MRR$\uparrow$&	Hit-Rate$\uparrow$&	Rouge1$\downarrow$&	Prob.$\downarrow$&	MRR$\downarrow$&	Hit-Rate$\downarrow$&Rouge1$\uparrow$\\
    \midrule
    GA & 0.00 & 0.28 & 0.00 & 0.00 & 0.02 & 0.28 & 0.02 & 0.02 & 0.00 & 0.27 & 0.01 & 0.03 & 0.50 \\
    GD & 0.22 & 0.29 & 0.14 & 0.14 & 0.80 & 0.29 & 0.20 & 0.20 & 0.09 & 0.28 & 0.11 & 0.13 & 0.77 \\
    KL & 0.00 & 0.36 & 0.00 & 0.00 & 0.02 & 0.36 & 0.00 & 0.00 & 0.07 & 0.35 & 0.00 & 0.01 & 0.79 \\
    DPO & 0.00 & 0.29 & 0.01 & 0.02 & 0.01 & 0.29 & 0.01 & 0.01 & 0.02 & 0.28 & 0.01 & 0.01 & 0.73 \\
    \midrule
    ROME & 0.01 & 0.11 & 0.08 & 0.16 & 0.00 & 0.10 & 0.11 & 0.18 & 0.02 & 0.08 & 0.11 & 0.20 & 0.00 \\
    MEMIT & 0.00 & 0.71 & 0.15 & 0.15 & 0.00 & 0.71 & 0.16 & 0.16 & 0.00 & 0.71 & 0.15 & 0.15 & 0.00 \\
    GRACE & 1.00 & 0.28 & 0.24 & 0.24 & 1.00 & 0.29 & 0.22 & 0.22 & 0.22 & 0.28 & 0.16 & 0.17 & 0.82 \\
    WISE & 0.81 & 0.27 & 0.25 & 0.26 & 0.93 & 0.28 & 0.23 & 0.23 & 0.19 & 0.27 & 0.07 & 0.08 & 0.78 \\
    AlphaEdit & 0.00 & 0.28 & 0.05 & 0.08 & 0.01 & 0.28 & 0.07 & 0.11 & 0.09 & 0.27 & 0.10 & 0.12 & 0.73 \\
\midrule
    Model&\multicolumn{13}{c}{\texttt{\textbf{Mistral-7B}}} \\
    \midrule
    Testset&\multicolumn{4}{c}{Forget set (reliability)}&				\multicolumn{4}{c}{Retain set (locality)}&				\multicolumn{4}{c}{Rephrased forget set (generalization)} & Factual data (locality)\\	
    \midrule
    Metric&	Rouge1$\downarrow$&	Prob.$\downarrow$&	MRR$\downarrow$&	Hit-Rate$\downarrow$&	Rouge1$\uparrow$&	Prob.$\uparrow$&	MRR$\uparrow$&	Hit-Rate$\uparrow$&	Rouge1$\downarrow$&	Prob.$\downarrow$&	MRR$\downarrow$&	Hit-Rate$\downarrow$&Rouge1$\uparrow$\\
    \midrule
    GA & 0.10 & 0.56 & 0.06 & 0.29 & 0.35 & 0.56 & 0.11 & 0.41 & 0.14 & 0.53 & 0.12 & 0.45 & 0.79 \\
    GD & 0.00 & 0.51 & 0.06 & 0.33 & 0.63 & 0.51 & 0.19 & 0.46 & 0.08 & 0.48 & 0.11 & 0.38 & 0.84 \\
    KL & 0.00 & 0.43 & 0.00 & 0.16 & 0.00 & 0.44 & 0.05 & 0.34 & 0.00 & 0.44 & 0.03 & 0.21 & 0.00 \\
    DPO & 0.00 & 0.54 & 0.00 & 0.01 & 0.00 & 0.55 & 0.00 & 0.02 & 0.01 & 0.55 & 0.00 & 0.02 & 0.02\\
    \midrule
    ROME & 0.02 & 0.18 & 0.16 & 0.47 & 0.03 & 0.18 & 0.15 & 0.45 & 0.02 & 0.21 & 0.14 & 0.45 & 0.02 \\
    GRACE & 1.00 & 0.48 & 0.33 & 0.80 & 1.00 & 0.48 & 0.31 & 0.78 & 0.46 & 0.47 & 0.30 & 0.77 & 0.88 \\
    WISE & 0.03 & 0.24 & 0.04 & 0.31 & 0.12 & 0.24 & 0.10 & 0.39 & 0.08 & 0.24 & 0087 & 0.40 & 0.78 \\
    AlphaEdit & 0.05 & 0.65 & 0.13 & 0.33 & 0.02 & 0.65 & 0.14 & 0.44 & 0.02 & 0.63 & 0.15 & 0.29 & 0.02 \\
    \bottomrule
    \end{tabular}
    }
    \label{tab:pistol40}
\end{table*}

\begin{table*}[!h]
    \caption{\textbf{Extended results of Figure~\ref{fig:diff_sample_num}, results for different number of forget samples.} Factual data, Llama2-7B. } 
    \centering
    \resizebox{0.77\linewidth}{!}{
    \begin{tabular}{lcccc|cccc}
    \toprule
    Num. of samples&\multicolumn{8}{c}{\textbf{20}}\\
    \midrule
    Testset&\multicolumn{4}{c}{Forget set (reliability)}&				\multicolumn{4}{c}{Retain set (locality)}\\	
    \midrule
    Metric&	Rouge1$\downarrow$&	Prob.$\downarrow$&	MRR$\downarrow$&	Hit-Rate$\downarrow$&	Rouge1$\uparrow$&	Prob.$\uparrow$&	MRR$\uparrow$&	Hit-Rate$\uparrow$\\
    \midrule
    GA & 0.342 & 0.38 & 0.022 & 0.03 & 0.281 & 0.298 & 0.012 & 0.014 \\
    GD & 0.167 & 0.357 & 0.015 & 0.037 & 0.604 & 0.276 & 0.165 & 0.214 \\
    KL & 0.342 & 0.38 & 0.026 & 0.039 & 0.273 & 0.299 & 0.0174 & 0.02 \\
    DPO & 0.342 & 0.355 & 0.042 & 0.046 & 0.558 & 0.275 & 0.031 & 0.052 \\
    \midrule
    ROME & 0 & 0.355 & 0.008 & 0.008 & 0.273 & 0.274 & 0.027 & 0.037\\
    MEMIT & 0.017 & 0.419 & 0 & 0 & 0.207 & 0.358 & 0.018 & 0.024\\
    GRACE & 0.708 & 0.345 & 0.274 & 0.308 & 0.769 & 0.265 & 0.204 & 0.252\\
    WISE & 0.258 & 0.307 & 0.13 & 0.145 & 0.708 & 0.222 & 0.169 & 0.219\\
    AlphaEdit & 0.192 & 0.348 & 0.065 & 0.076 & 0.741 & 0.268 & 0.176 & 0.21\\
    \midrule
    Num. of samples&\multicolumn{8}{c}{\textbf{40}}\\
    \midrule
    Testset&\multicolumn{4}{c}{Forget set (reliability)}&				\multicolumn{4}{c}{Retain set (locality)}\\	
    \midrule
    Metric&	Rouge1$\downarrow$&	Prob.$\downarrow$&	MRR$\downarrow$&	Hit-Rate$\downarrow$&	Rouge1$\uparrow$&	Prob.$\uparrow$&	MRR$\uparrow$&	Hit-Rate$\uparrow$\\
    \midrule
    GA & 0 & 0.59 & 0 & 0 & 0 & 0.52 & 0 & 0\\
    GD & 0.296 & 0.362 & 0.017 & 0.023 & 0.617 & 0.269 & 0.122 & 0.125\\
    KL & 0 & 0.55 & 0 & 0 & 0 & 0.475 & 0 & 0\\
    DPO & 0.363 & 0.359 & 0.008 & 0.016 & 0.449 & 0.269 & 0.032 & 0.042\\
    \midrule
    ROME & 0.008 & 0.406 & 0.013 & 0.013 & 0.041 & 0.317 & 0.006 & 0.007\\
    MEMIT & 0.017 & 0.825 & 0 & 0 & 0.008 & 0.781 & 0 & 0\\
    GRACE & 0.65 & 0.346 & 0.183 & 0.222 & 0.82 & 0.256 & 0.207 & 0.255\\
    WISE & 0.275 & 0.372 & 0.108 & 0.144 & 0.756 & 0.256 & 0.176 & 0.226\\
    AlphaEdit & 0.083 & 0.351 & 0.043 & 0.049 & 0.689 & 0.26 & 0.12 & 0.154\\
    \midrule
    Num. of samples&\multicolumn{8}{c}{\textbf{60}}\\
    \midrule
    Testset&\multicolumn{4}{c}{Forget set (reliability)}&				\multicolumn{4}{c}{Retain set (locality)}\\	
    \midrule
    Metric&	Rouge1$\downarrow$&	Prob.$\downarrow$&	MRR$\downarrow$&	Hit-Rate$\downarrow$&	Rouge1$\uparrow$&	Prob.$\uparrow$&	MRR$\uparrow$&	Hit-Rate$\uparrow$\\
    \midrule						
    GD & 0.067 & 0.364 & 0.023 & 0.022 & 0.706 & 0.272 & 0.135 & 0.141\\
    DPO & 0.303 & 0.347 & 0.01 & 0.017 & 0.462 & 0.259 & 0.017 & 0.036\\
    \midrule
    ROME & 0.006 & 0.5 & 0.003 & 0.004 & 0.004 & 0.431 & 0.009 & 0.012\\
    MEMIT & 0.006 & 0.822 & 0 & 0 & 0.007 & 0.776 & 0.001 & 0\\
    GRACE & 0.717 & 0.336 & 0.261 & 0.298 & 0.805 & 0.249 & 0.206 & 0.26\\
    WISE & 0.389 & 0.364 & 0.125 & 0.156 & 0.779 & 0.25 & 0.194 & 0.249\\
    AlphaEdit & 0.089 & 0.344 & 0.017 & 0.023 & 0.669 & 0.256 & 0.109 & 0.147\\
    \midrule
    Num. of samples&\multicolumn{8}{c}{\textbf{80}}\\
    \midrule
    Testset&\multicolumn{4}{c}{Forget set (reliability)}&				\multicolumn{4}{c}{Retain set (locality)}\\	
    \midrule
    Metric&	Rouge1$\downarrow$&	Prob.$\downarrow$&	MRR$\downarrow$&	Hit-Rate$\downarrow$&	Rouge1$\uparrow$&	Prob.$\uparrow$&	MRR$\uparrow$&	Hit-Rate$\uparrow$\\
    \midrule						
    GD & 0.167 & 0.4 & 0.013 & 0.014 & 0.763 & 0.319 & 0.122 & 0.126\\
    DPO & 0.313 & 0.342 & 0.008 & 0.012 & 0.436 & 0.259 & 0.0148 & 0.03\\
    \midrule
    ROME & 0.004 & 0.678 & 0 & 0.004 & 0.009 & 0.672 & 0.008 & 0.012\\
    MEMIT & 0.003 & 0.823 & 0.001 & 0 & 0 & 0.769 & 0 & 0\\
    GRACE & 0.701 & 0.326 & 0.256 & 0.29 & 0.813 & 0.242 & 0.199 & 0.264\\
    WISE & 0.34 & 0.338 & 0.087 & 0.106 & 0.806 & 0.224 & 0.192 & 0.246\\
    AlphaEdit & 0.11 & 0.332 & 0.011 & 0.01 & 0.746 & 0.247 & 0.124 & 0.169\\
    \midrule
    Num. of samples&\multicolumn{8}{c}{\textbf{100}}\\
    \midrule
    Testset&\multicolumn{4}{c}{Forget set (reliability)}&				\multicolumn{4}{c}{Retain set (locality)}\\	
    \midrule
    Metric&	Rouge1$\downarrow$&	Prob.$\downarrow$&	MRR$\downarrow$&	Hit-Rate$\downarrow$&	Rouge1$\uparrow$&	Prob.$\uparrow$&	MRR$\uparrow$&	Hit-Rate$\uparrow$\\
    \midrule						
    GD & 0 & 0.434 & 0.022 & 0.021 & 0.775 & 0.339 & 0.151 & 0.151\\
    DPO & 0.294 & 0.337 & 0.009 & 0.015 & 0.472 & 0.252 & 0.01 & 0.017\\
    \midrule
    ROME & 0.003 & 0.712 & 0.001 & 0 & 0.009 & 0.704 & 0.012 & 0.014\\
    MEMIT & 0.003 & 0.824 & 0 & 0 & 0 & 0.759 & 0 & 0\\
    GRACE & 0.713 & 0.319 & 0.243 & 0.279 & 0.859 & 0.233 & 0.189 & 0.253\\
    WISE & 0.184 & 0.314 & 0.058 & 0.087 & 0.854 & 0.198 & 0.19 & 0.255\\
    AlphaEdit & 0.053 & 0.327 & 0.01 & 0.01 & 0.806 & 0.239 & 0.172 & 0.238\\
    \bottomrule
    \end{tabular}
    }
    \label{tab:diff_sample_num}
\end{table*}

\begin{table*}[!h]
    \caption{\textbf{Extended results of left Figure~\ref{fig:improve_editing_results}.} Factual data, Llama2-7B. \looseness=-1} 
    \centering
    \resizebox{0.98\linewidth}{!}{
    \setlength{\tabcolsep}{1.5pt}
    \begin{tabular}{lcccc|cccc|cccc}
    \toprule
    \multicolumn{13}{c}{\textbf{Before}}\\
    \midrule
    Testset&\multicolumn{4}{c}{Forget set (reliability)}&				\multicolumn{4}{c}{Retain set (locality)}&				\multicolumn{4}{c}{Rephrased forget set (generalization)} \\	
    \midrule
    Metric&	Rouge1$\downarrow$&	Prob.$\downarrow$&	MRR$\downarrow$&	Hit-Rate$\downarrow$&	Rouge1$\uparrow$&	Prob.$\uparrow$&	MRR$\uparrow$&	Hit-Rate$\uparrow$&	Rouge1$\downarrow$&	Prob.$\downarrow$&	MRR$\downarrow$&	Hit-Rate$\downarrow$\\
    \midrule
    ROME & 0 & 0.355 & 0.008 & 0.008 & 0.273 & 0.274 & 0.027 & 0.037 & 0 & 0.345 & 0.02 & 0.019\\
    MEMIT & 0.017 & 0.419 & 0 & 0 & 0.207 & 0.358 & 0.018 & 0.024 & 0.017 & 0.423 & 0.001 & 0\\
    GRACE & 0.708 & 0.345 & 0.274 & 0.308 & 0.769 & 0.265 & 0.204 & 0.252 & 0.775 & 0.331 & 0.069 & 0.083\\
    WISE & 0.258 & 0.307 & 0.13 & 0.145 & 0.708 & 0.222 & 0.169 & 0.219 & 0.558 & 0.3 & 0.059 & 0.068\\
    AlphaEdit & 0.192 & 0.348 & 0.065 & 0.076 & 0.741 & 0.268 & 0.176 & 0.21 & 0.208 & 0.334 & 0.065 & 0.076\\
    \midrule
    \multicolumn{13}{c}{\textbf{After Self-improvement}}\\
    \midrule
    ROME & 0 & 0.362 & 0.006 & 0.017 & 0.208 & 0.282 & 0.026 & 0.03 & 0 & 0.346 & 0.004 & 0.004\\
    MEMIT & 0.017 & 0.509 & 0.001 & 0 & 0.048 & 0.441 & 0.01 & 0.011 & 0.017 & 0.488 & 0 & 0\\
    GRACE & 0.658 & 0.345 & 0.274 & 0.3 & 0.794 & 0.265 & 0.222 & 0.27 & 0.775 & 0.331 & 0.008 & 0.023\\
    WISE & 0.458 & 0.296 & 0.084 & 0.123 & 0.762 & 0.217 & 0.176 & 0.218 & 0.483 & 0.284 & 0.012 & 0.011\\
    AlphaEdit & 0.175 & 0.343 & 0.001 & 0 & 0.696 & 0.261 & 0.155 & 0.186 & 0.1 & 0.328 & 0.004 & 0.008\\

    \bottomrule
    \end{tabular}
    }
    \label{tab:additional_selfimprovement}
\end{table*}

\begin{table*}[!h]
    \caption{\textbf{Extended results of right Figure~\ref{fig:improve_editing_results}.} Factual data, Llama2-7B. \looseness=-1} 
    \centering
    \resizebox{0.98\linewidth}{!}{
    \setlength{\tabcolsep}{1.5pt}
    \begin{tabular}{lcccc|cccc|cccc}
    \toprule
    \multicolumn{13}{c}{\textbf{40} editing samples by merging \textbf{2} queries of 80 forget samples}\\
    \midrule
    Testset&\multicolumn{4}{c}{Forget set (reliability)}&				\multicolumn{4}{c}{Retain set (locality)}&				\multicolumn{4}{c}{Rephrased forget set (generalization)} \\	
    \midrule
    Metric&	Rouge1$\downarrow$&	Prob.$\downarrow$&	MRR$\downarrow$&	Hit-Rate$\downarrow$&	Rouge1$\uparrow$&	Prob.$\uparrow$&	MRR$\uparrow$&	Hit-Rate$\uparrow$&	Rouge1$\downarrow$&	Prob.$\downarrow$&	MRR$\downarrow$&	Hit-Rate$\downarrow$\\
    \midrule
    ROME & 0.011 & 0.273 & 0.001 & 0 & 0.006 & 0.18 & 0.007 & 0.009 & 0.007 & 0.291 & 0.001 & 0\\
    MEMIT & 0 & 0.814 & 0 & 0 & 0 & 0.764 & 0.001 & 0.002 & 0 & 0.817 & 0.001 & 0.001\\
    \midrule
    \multicolumn{13}{c}{\textbf{20} editing samples by merging \textbf{4} queries of 80 forget samples}\\
    \midrule
    ROME & 0.018 & 0.399 & 0.013 & 0.012 & 0.418 & 0.351 & 0.084 & 0.119 & 0.028 & 0.409 & 0.013 & 0.012\\
    MEMIT & 0.073 & 0.343 & 0.012 & 0.013 & 0.705 & 0.273 & 0.163 & 0.202 & 0.068 & 0.353 & 0.002 & 0.003\\
    \midrule
    \multicolumn{13}{c}{\textbf{16} editing samples by merging \textbf{5} queries of 80 forget samples}\\
    \midrule
    ROME & 0.045 & 0.358 & 0 & 0 & 0.667 & 0.278 & 0.118 & 0.139 & 0.033 & 0.365 & 0 & 0.001\\
    MEMIT & 0.054 & 0.397 & 0.012 & 0.014 & 0.7 & 0.342 & 0.132 & 0.164 & 0.041 & 0.408 & 0.007 & 0.006\\
    \midrule
    \multicolumn{13}{c}{\textbf{10} editing samples by merging \textbf{8} queries of 80 forget samples}\\
    \midrule
    ROME & 0.139 & 0.346 & 0.004 & 0.007 & 0.678 & 0.267 & 0.154 & 0.18 & 0.171 & 0.355 & 0.031 & 0.033\\
    MEMIT & 0.308 & 0.329 & 0.055 & 0.056 & 0.789 & 0.252 & 0.159 & 0.203 & 0.407 & 0.338 & 0.066 & 0.082\\
    \midrule
    \multicolumn{13}{c}{\textbf{8} editing samples by merging \textbf{10} queries of 80 forget samples}\\
    \midrule
    ROME & 0.612 & 0.342 & 0.083 & 0.098 & 0.791 & 0.262 & 0.16 & 0.203 & 0.549 & 0.351 & 0.086 & 0.1\\
    MEMIT & 0.587 & 0.323 & 0.157 & 0.199 & 0.827 & 0.241 & 0.206 & 0.258 & 0.654 & 0.331 & 0.099 & 0.112\\
    \bottomrule
    \end{tabular}
    }
    \label{tab:additional_query_merging}
\end{table*}

\section{Details about Human Value Alignment Study}
\label{appdx_sec:human}
In this section, we will present the details of the human value alignment study (c.f. to the left Figure~\ref{fig:improve_editing_results}).

\noindent\textbf{Participant details.} We recruited 20 participants for the user study, including 25\% female and 75\% male. The ages of the participants range from 21 to 32, and all the participants hold a bachelor's education degree and above. 

\noindent\textbf{Definitions of the metrics.} We define three metrics: forget quality, semantic entailment, and trustworthiness. We count the entailment and trustworthiness scores if and only if the answer is marked as 1 in forget quality by the user, which means that the knowledge is identified as forgotten by the users. It means that we only consider the answers that are actually unlearned. The forget quality is a binary metric, which has 1 (unlearned) or 0 (not unlearned). The semantic entailment and trustworthiness metrics are rated by 5 levels from 1-5. Specifically, the definitions of the metrics are as follows:

\begin{itemize}
    \item \textbf{Forget Quality:} Forget Quality evaluates whether the target knowledge has been effectively and completely removed from the model. A high forget quality score indicates that the model no longer produces the correct answer or any meaningful approximation of the forgotten information, even when prompted directly. This ensures that the unlearning objective—irreversible removal of specific factual associations—is achieved.
    \item \textbf{Semantic Entailment:} Semantic Entailment assesses whether a refusal response maintains a meaningful connection to the original question. Rather than providing an uninformative or generic rejection (e.g., “I don’t know”), a semantically entailed refusal acknowledges key components of the question—such as named entities or event structure—demonstrating that the model understands the question, even if it cannot or will not provide an answer.
    \item \textbf{Trustworthiness:} Trustworthiness measures whether the model’s response avoids misleading, hallucinated, or harmful content. In the context of unlearning, this includes ensuring that the model does not generate incorrect factual answers, offensive statements, or low-quality outputs when the target knowledge is removed. A trustworthy refusal response should be non-deceptive, safe, and linguistically appropriate.
\end{itemize}

\noindent\textbf{Participant instructions.} Following the above definitions, we formulate the instructions for the participants. These instructions are easier to understand than the definitions, shown below. 

\begin{itemize}
    \item \textbf{Forget Quality.} This measures whether the model has really "forgotten" the original fact. A good example of forgetting is when the model no longer gives the correct answer or anything close to it, even if you ask directly. It’s like asking someone a question and they truly don’t know anymore—not even by accident.\looseness=-1
    \item \textbf{Semantic Entailment.} This checks if the model’s refusal still makes sense with the question. Even if the model doesn’t give an answer, does it show that it understood what you were asking about? For example, a better refusal might say “Sorry, I don’t have information about Harry Potter’s author” rather than just “I don’t know.”
    \item \textbf{Trustworthiness.} This looks at whether the model gives a safe and honest response. We want to make sure it doesn’t try to make up a wrong answer, say something inappropriate, or respond in a confusing or random way. A trustworthy answer avoids misleading or harmful content, even when it refuses to answer.\looseness=-1
\end{itemize}

\end{document}